\def\year{2019}\relax
\documentclass[letterpaper]{article} 
\pdfoutput=1
\usepackage{aaai19}  
\usepackage{times}  
\usepackage{helvet}  
\usepackage{courier}  
\usepackage{url}  
\usepackage{graphicx}  
\graphicspath{{./imgs/}}

\usepackage{bm}
\usepackage{amsmath}
\usepackage{amssymb}
\usepackage{subfig}
\usepackage{enumerate}
\usepackage{bbm}
\usepackage{multirow}

\DeclareMathOperator{\Tr}{tr}
\title{Hyperprior Induced Unsupervised Disentanglement of Latent Representations}
\author{Abdul Fatir Ansari \and Harold Soh\\
	Department of Computer Science, National University of Singapore\\
	\{afatir, harold\}@comp.nus.edu.sg}
\begin{document}
	\maketitle
	\begin{abstract}
		We address the problem of unsupervised disentanglement of latent representations learnt via deep generative models. In contrast to current approaches that operate on the evidence lower bound (ELBO), we argue that statistical independence in the latent space of VAEs can be enforced in a principled hierarchical Bayesian manner. To this effect, we augment the standard VAE with an inverse-Wishart (IW) prior on the covariance matrix of the latent code. By tuning the IW parameters, we are able to encourage (or discourage) independence in the learnt latent dimensions. Extensive experimental results on a range of datasets (2DShapes, 3DChairs, 3DFaces and CelebA) show our approach to outperform the $\beta$-VAE and is competitive with the state-of-the-art FactorVAE. Our approach achieves significantly better disentanglement and reconstruction on a new dataset (CorrelatedEllipses) which introduces correlations between the factors of variation. 
	\end{abstract}

\section{Introduction}

Learning semantically interpretable representations of data remains an important open problem in artificial intelligence. In particular, there has been considerable attention on learning \textit{disentangled} representations---equivariant codes that exhibit predictable changes in a \emph{single} associated dimension when a factor of variation is altered~\cite{Bengio2013RepresentationLA}. Disentangled representations are beneficial for a variety of tasks including exploratory data analysis (EDA), transfer learning, and generative modeling. For example, one may seek to change a single aspect of a generated face (e.g., lighting, orientation, or hair color). With an appropriate disentangled representation, only one dimension of the latent code needs to be modified to obtain the required change. By analogy to inverse graphics, the disentangled representation can be regarded as independent parameters fed to a rendering engine to synthesize an image.

In this work, we focus on pure \emph{unsupervised} learning of disentangled representations with deep generative models. Alternative (semi-) supervised approaches have been explored in recent work~\cite{Reed2014LearningTD,Kulkarni2015DeepCI,Siddharth2017LearningDR,Mathieu2016DisentanglingFO}, but these methods require labels that can be costly to obtain. Moreover, in certain applications such as EDA, the factors of variation are unknown and precisely the information we seek to uncover. 

In the unsupervised setting, a prior notion of disentanglement is required: we adopt a current standard assumption that the data is generated from a fixed number of \emph{statistically independent} factors. A popular generative model under this assumption is the $\beta$-VAE~\cite{BetaVAE_Matthey2016vaeLB}, which learns good disentangled representations whilst being easy to train. However, the mechanism employed to encourage disentanglement---by increasing the weight on the KL divergence between the variational posterior and prior---sacrifices reconstruction fidelity. 

Recognizing this deficiency, very recent models---the FactorVAE~\cite{FactorVAE_Kim2018DisentanglingBF} and $\beta$-TCVAE~\cite{Chen2018IsolatingSO}---have improved upon the $\beta$-VAE by augmenting the VAE loss with an extra penalty term that encourages independence in the latent codes. Although this penalty term is well-motivated via \emph{total correlation}, the drive to maximize statistical independence in this manner may not be robust to factor correlations in the data that exist due to biased sampling. Furthermore, the need to add additional weighted terms to the variational lower bound is unsatisfying from a probabilistic modeling perspective; it points to an inadequacy in the underlying model formulation.

This paper takes a step back and asks whether trading-off reconstruction and disentanglement can be handled in a more principled fashion. Rather than introducing additional terms or weights to the VAE evidence lower bound (ELBO), we focus instead on the generative model and specifically, its \emph{latent representation prior} $p(\mathbf{z})$. In the standard VAE, $p(\mathbf{z})$ is a standard multivariate Gaussian $p(\mathbf{z}) = \mathcal{N}(\mathbf{0}, \mathbf{\Sigma})$ where $\mathbf{\Sigma} = \mathbf{I}$. By introducing a suitable \emph{hyperprior} on the covariance matrix---e.g., the inverse-Wishart (IW) used in this study---one imagines that we can encourage (or discourage) independence in the learnt latent dimensions via the hyperprior's  parameters. When trained via variational inference using an approximate distribution $q(\mathbf{z},\mathbf{\Sigma})$, the hyperprior's effect naturally manifests as additional terms in the ELBO, rather than having to be inserted post-hoc as in previous studies. This approach is very natural from a Bayesian perspective, but surprisingly, has yet to be explored in the literature. 

Unlike previous work, our model formulation entails learning a full covariance matrix $\mathbf{\Sigma}$; this allows the model to capture possible correlations in the dataset, but requires additional treatment to ensure stable training. We employ a \emph{structured variational posterior} and present approximation techniques to enable efficient and stable inference. We term the resulting model and inference scheme as the Covariance Hyperprior VAE (CHyVAE). 

Experiments on a range of image datasets---2DShapes, CelebA, 3DFaces, and 3DChairs---show that CHyVAE outperforms $\beta$-VAE both in terms of disentanglement and reconstruction error and is competitive with the state-of-the-art FactorVAE. We also compare the three models on a novel dataset (CorrelatedEllipses) which introduces strong correlations between the factors of variation. Here, CHyVAE outperforms both $\beta$-VAE and FactorVAE by a significant margin. These results indicate that disentanglement and reconstruction can be traded-off in an alternative manner, i.e., at the model specification level, compared to existing approaches that operate on the ELBO.

In summary, this paper makes the following key contributions:
\begin{itemize}
	\item A hierarchical Bayesian approach for learning disentangled latent space representations in an unsupervised manner;
	\item A specific generative model with an inverse-Wishart hyperprior and an efficient inference scheme with a structured variational posterior, which results in the CHyVAE;
	\item Extensive empirical results and analyses comparing CHyVAE to $\beta$-VAE and the FactorVAE on a range of datasets, which show that disentanglement can be achieved without resorting to ``ELBO surgery''~\cite{hoffman2016elbo,FactorVAE_Kim2018DisentanglingBF,Chen2018IsolatingSO}.
\end{itemize}

\section{Generation and Disentanglement with the Variational Autoencoder}
\label{sec:background}

To begin, we give a brief overview of the Variational Autoencoder (VAE)~\cite{VAE_Kingma2013AutoEncodingVB}\footnote{We refer readers wanting more detail on the VAE and an alternative derivation to related work~\cite{Doersch2016,VAE_Kingma2013AutoEncodingVB}.} and the alterations used to encourage disentanglement. We first consider the standard generative scheme, where our objective is to find parameters $\theta$ that maximize the expected log probability of the dataset under the data distribution,
\begin{align}
\text{argmax}_\theta \mathbb{E}_\text{data}[\log p_\theta(\mathbf{x})]
\label{eq:maxlogpd}
\end{align}
and $\mathbf{x} \in \mathcal{X}$ is an observed data item of interest. For real world data, $p_\theta(\mathbf{x})$ may be highly complex and non-trivial to generate samples from. Furthermore, in unsupervised learning, we may wish to obtain representations of $\mathbf{x}$ that are more amenable to downstream analysis. One approach for achieving these aims is to further specify the log-distribution within Eq. (\ref{eq:maxlogpd}),
\begin{align}
\log p_\theta(\mathbf{x}) = \log \mathbb{E}_{p(\mathbf{z})}[p_\theta(\mathbf{x}|\mathbf{z})] 
\end{align}
where we introduce the conditional distribution $p_\theta(\mathbf{x}|\mathbf{z})$ and variables $\mathbf{z} \in \mathcal{Z}$ with prior $p(\mathbf{z})$. Intuitively, each $\mathbf{z}$ is a latent representation or code associated with $\mathbf{x}$. By choosing an appropriate condition and setting the prior $p(\mathbf{z})$ to be a simple distribution, e.g., $\mathcal{N}(\mathbf{0}, \mathbf{I})$, we can easily generate $\mathbf{x}$ by sampling from $p(\mathbf{z})$. In addition, the diagonal covariance $\mathbf{I}$ indicates a prior expectation that underlying data representation comprises statistically independent Gaussians (one for each latent dimension), and is therefore disentangled.

Computing $\log p_\theta(\mathbf{x})$ requires marginalizing out the latent variables $\mathbf{z}$ which is generally intractable, e.g., when $p_\theta(\mathbf{x}|\mathbf{z}) = p(\mathbf{x}|f(\mathbf{z}))$ and $f$ is a nonlinear neural network. To perform approximate inference, the VAE employs a \emph{recognition} or \emph{inference} \emph{model}, $q_\phi(\mathbf{z}|\mathbf{x})$, 
and maximizes the variational or evidence lower bound (ELBO), 

\begin{align*}
\mathcal{L}^{\mathrm{VAE}}_{\mathrm{ELBO}} = \frac{1}{N}\sum_{n=1}^N \mathbb{E}_{q_\phi}[\log p_\theta(\mathbf{x}|\mathbf{z})] - D_{\mathrm{KL}}(q_\phi(\mathbf{z}|\mathbf{x})||p(\mathbf{z}))
\end{align*}
The above can be seen as the expectation of the data likelihood under the inference model with a KL divergence term that measures how different $q_\phi(\mathbf{z}|\mathbf{x})$ is from the prior $p(\mathbf{z})$.

\subsection{Encouraging Disentanglement in VAEs}
In our context, the second term in the ELBO is of particular interest: when $p(\mathbf{z})$ is intentionally chosen to factorize across the dimensions, minimizing the KL divergence encourages independence in the learned latent representations. The $\beta$-VAE \cite{BetaVAE_Matthey2016vaeLB} takes advantage of this observation and encourages disentanglement by emphasizing the KL divergence with a weight $\beta$:
\begin{align*}
\mathcal{L}^{\beta} = \frac{1}{N}\sum_{n=1}^N \mathbb{E}_{q}[\log p(\mathbf{x}|\mathbf{z})] - \beta D_{\mathrm{KL}}[q(\mathbf{z}|\mathbf{x})\|p(\mathbf{z})]
\end{align*}
With larger $\beta$ and a factorized prior, maximizing this modified objective favors latent representations possessing greater independence across the dimensions. 

However, disentanglement gains in the $\beta$-VAE are often off-set by a decrease in reconstruction performance. Recent work~\cite{FactorVAE_Kim2018DisentanglingBF,Chen2018IsolatingSO} has argued that increasing $\beta$ has the undesirable side-effect of inadvertently penalizing the mutual information between $\mathbf{x}$ and $\mathbf{z}$. This can be seen by decomposing the KL divergence term:
\begin{align*}
\mathbb{E}_\mathrm{data}[D_{\mathrm{KL}}[q(\mathbf{z}|\mathbf{x})\|p(\mathbf{z}))]] = I(\mathbf{x}; \mathbf{z}) + D_{\mathrm{KL}}[q(\mathbf{z})\|p(\mathbf{z})]
\end{align*}
where $I(\mathbf{x}; \mathbf{z})$ is the mutual information between $\mathbf{x}$ and $\mathbf{z}$~\cite{hoffman2016elbo}. Penalizing the first term decreases the informativeness of $\mathbf{z}$ about $\mathbf{x}$ and hence, reduces reconstruction quality. To overcome this problem, both \citeauthor{FactorVAE_Kim2018DisentanglingBF} and \citeauthor{Chen2018IsolatingSO} optimize an augmented objective:
\begin{align}
\mathcal{L}^{\gamma} = \mathcal{L}^{\mathrm{VAE}}_{\mathrm{ELBO}} -\gamma D_{\mathrm{KL}}\left[q(\mathbf{z})\|\prod_{i}q(z_i)\right]
\label{eq:lgamma}
\end{align}
where $ D_{\mathrm{KL}}[q(\mathbf{z})\|\prod_{i}q(z_i)]$ is the \textit{total correlation} (TC). Computing the TC is generally intractable and has to be approximated; the FactorVAE~\cite{FactorVAE_Kim2018DisentanglingBF} estimates TC by means of a discriminator using the density-ratio trick, where else $\beta$-TCVAE~\cite{Chen2018IsolatingSO} uses a minibatch-based alternative. Empirical results show that optimizing $\mathcal{L}^{\gamma}$ leads to better reconstruction with similar disentanglement compared to the $\beta$-VAE.

\section{A Hyperprior Approach for Learning Disentangled Representations}

We observed in the previous section that current state-of-the-art methods attempt to procure disentanglement by augmenting the ELBO. In this section, we describe an alternative approach by further expanding upon the log-distribution within Eq. (\ref{eq:maxlogpd}). At a high-level, we desire a means to ``regularize'' the latent codes towards a disentangled form, yet preserve sufficient \emph{flexibility} to achieve good reconstruction. A natural Bayesian approach to achieve these aims is to place a hyperprior $p(\bf\Sigma)$ on the covariance parameter of  $p(\mathbf{z}|\bf\Sigma)$:
\begin{align}
\log p_\theta(\mathbf{x}) = \log \mathbb{E}_{ p(\mathbf{z} | \mathbf{\Sigma} ) p(\bf{\Sigma} ) }[p(\mathbf{x}|\mathbf{z})].
\label{eq:logpdata_hyp}
\end{align}

Our proposed model denotes a modified generative process relative to the standard VAE. In particular, an observed sample $\mathbf{x} \in \mathbb{R}^{D}$ is generated by first sampling a covariance matrix $\bm{\Sigma} \sim p(\bm\Sigma)$, followed by the latent representation $\mathbf{z} \sim p(\bf{z} | \bm\Sigma)$ and finally, $\mathbf{x} \sim p_\theta(\mathbf{x}|\mathbf{z})$. As such, the joint probability factorizes as
\begin{align}
p(\mathbf{x},\mathbf{z},\bm{\Sigma}) = p(\mathbf{x}|\mathbf{z})p(\mathbf{z}|\bm{\Sigma})p(\bm{\Sigma})
\label{eq:jointfactors}
\end{align}
since $\mathbf{x}$ and $\bm{\Sigma}$ are independent conditioned on $\mathbf{z}$. Notice that $\bm\Sigma$ is no longer constrained to be simply $\bf I$, but disentanglement can be encouraged in a straight-forward manner simply by placing greater weight on independence between the latent dimensions. In other words, tuning the strength or \emph{informativeness} of the hyperprior would then allow us to naturally vary of level of disentanglement desired. By allowing some deviation from strict independence, the model recognizes that individual latent representations \emph{may} have correlated factors of variation; this is potentially a more accurate reflection of real world data where different sub-populations (e.g., dog breeds) often have correlated factors of variation (e.g., color, size). 

\subsubsection{ELBO Decomposition}
Akin to the VAE, inference can be achieved via variational approximation. Let $q(\mathbf{z},\bm{\Sigma}|\mathbf{x})$ be the variational posterior distribution. Using Jensen's inequality, the log-likelihood can be written as
\begin{align}
\log p(\mathbf{x}) \geq \mathbb{E}_{q(\mathbf{z},\bm{\Sigma}|\mathbf{x})}\left[\log \frac{p(\mathbf{z},\bm{\Sigma},\mathbf{x})}{q(\mathbf{z},\bm{\Sigma}|\mathbf{x})}\right] = \mathcal{L}_{\mathrm{ELBO}}
\label{eq:elbo}
\end{align} 

Consider a structured variational distribution $q(\mathbf{z},\bm{\Sigma}|\mathbf{x})=q(\mathbf{z}|\mathbf{x})q(\bm{\Sigma}|\mathbf{z})$
. Then, the ELBO in Eq. (\ref{eq:elbo}) decomposes into the following terms: 
\begin{align}
\nonumber\mathcal{L}_{\mathrm{ELBO}} = \underset{\textnormal{(average reconstruction)}}{\left[\frac{1}{N}\sum_{n=1}^N\mathbb{E}_{q(\mathbf{z}|\mathbf{x})}\left[\log p(\mathbf{x}|\mathbf{z})\right]\right]} - \underset{\textnormal{(index-code MI)}}{I(\mathbf{x};\mathbf{z})}\\
- \underset{\textnormal{(marginal KL to prior)}}{D_{\mathrm{KL}}(q(\mathbf{z})\|p(\mathbf{z}))} - \underset{\textnormal{(covariance penalty)}}{\mathbb{E}_{q(\mathbf{z})}\left[D_{\mathrm{KL}}(q(\bm{\Sigma}|\mathbf{z})\|p(\bm{\Sigma}|\mathbf{z}))\right]} \label{eq:elbosurgery}
\end{align}
We can combine the first three terms in Eq. (\ref{eq:elbosurgery}) to obtain the standard VAE ELBO~\cite{hoffman2016elbo}:
\begin{align}
\nonumber\mathcal{L}_{\mathrm{ELBO}} &= \mathcal{L}^{\mathrm{VAE}}_{\mathrm{ELBO}} - \mathbb{E}_{q(\mathbf{z})}\left[D_{\mathrm{KL}}(q(\bm{\Sigma}|\mathbf{z})\|p(\bm{\Sigma}|\mathbf{z}))\right]
\end{align}
revealing $\mathbb{E}_{q(\mathbf{z})}\left[D_{\mathrm{KL}}(q(\bm{\Sigma}|\mathbf{z})\|p(\bm{\Sigma}|\mathbf{z}))\right]$ as an additional quantity minimized in our model. Intuitively, this term matches the approximate covariance distribution to the covariance prior (across the latent code distribution). Hence, a factorized $p(\bm{\Sigma}|\mathbf{z})$ can encourage disentangled representations. A similar term was introduced directly into the ELBO in the DIP-VAE~\cite{Kumar2017VariationalIO}, which constrains the covariance matrix (averaged over the mini-batch) to be near $\mathbf{I}$. In contrast, our hyperprior approach enables this term to emerge naturally in the ELBO. 

\subsection{Model Specification and Inference}
In this subsection, we derive a specific model---the Covariance Hyperprior VAE (CHyVAE)---under the hyperprior framework outlined above. Specifically, we set a Gaussian prior over the latent code, and an inverse-Wishart prior over its covariance matrix. Note that the overarching framework is not constrained to these specific distributions, i.e., alternative hyperpriors and variational distributions can be used without significant changes to the overall methodology. 

\subsubsection{An Inverse-Wishart Hyperprior}

\begin{figure}
	\centering
	\includegraphics[width=0.9\columnwidth]{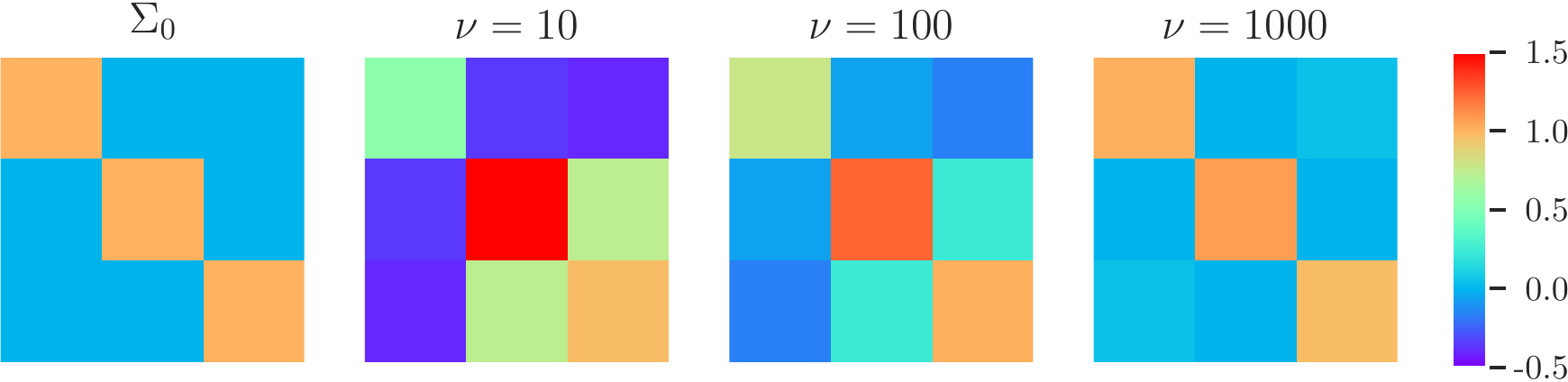}
	\caption{Desired matrix $\Sigma_0$ and a random sample from an inverse-Wishart distribution with $\nu=10$, $\nu=100$, and $\nu=1000$. As $\nu$ increases, random samples become closer to the desired matrix.} 
	\label{fig:invwishart}
\end{figure}

The inverse-Wishart distribution $\mathcal{W}_{p}^{-1}(\bm{\Psi}, \nu)$ is a popular distribution that has support over real-valued positive-definite matrices. It is parameterized by a positive-definite scale matrix $\bm{\Psi} \in \mathbb{R}^{p \times p}$ and degrees of freedom (DoF) $\nu > p - 1$. An inverse-Wishart distributed random matrix $\mathbf{X} \in \mathbb{R}^{p \times p}$ has probability density function
\begin{align*}
p(\mathbf{X}) = \frac{|\bm{\Psi}|^{\nu/2}}{2^{\nu p/2}\Gamma_p\left(\frac{\nu}{2}\right)} |\mathbf{X}|^{-(\nu + p + 1)/2} e^{-\frac{1}{2}\Tr(\bm{\Psi}\mathbf{X}^{-1})}
\end{align*}
where $\Gamma_p(.)$ is the multivariate gamma function. The mean of an inverse-Wishart random variable is given by $(\nu - p - 1)^{-1}\bm{\Psi}$. For a desired specification $\bm{\Sigma}_0$ of the covariance matrix, a reasonable choice of $\bm{\Psi}$ would be $(\nu - p - 1)\bm{\Sigma}_0$.  

With a desired covariance matrix $\mathbf{I}$, the DoF parameter $\nu$ can be varied to control the desired statistical independence (Fig. \ref{fig:invwishart}). Intuitively, $\nu$ can be regarded as pseudo-observations and thus, controls the strength/informativeness of the prior; high values ($\nu \gg p$) indicate a strong prior, while $\nu = p + 1$ is the least informative setting. 

\subsubsection{Approximate Inference}
In our model, we have
\begin{align}
p(\bm{\Sigma}) &= \mathcal{W}_p^{-1}(\bm{\Sigma}|\bm{\Psi}, \nu)\\
p(\mathbf{z}|\bm{\Sigma}) &= \mathcal{N}(\mathbf{z}|\bm{0}, \bm{\Sigma})
\end{align}
and maximizing the log probability of the data (Eq. \ref{eq:logpdata_hyp}) is intractable. Our initial approach was to completely employ a mean-field variational approximation, i.e., factorize $q(\mathbf{z},\bm{\Sigma}|\mathbf{x})$ into $q(\mathbf{z}|\mathbf{x})q(\bm{\Sigma}|\mathbf{x})$. This factorization can be realized using shared neural networks that output both the prior and the hyperprior parameters. While simple and tractable, our preliminary experiments with this factorized form were unsuccessful; training was unstable and results were poor. One potential reason is that the decoupling renders the hyperprior ineffective. 

An alternative approach is to factorize $q(\mathbf{z},\bm{\Sigma}|\mathbf{x})=q(\mathbf{z}|\mathbf{x},\bm{\Sigma})q(\bm{\Sigma}|\mathbf{x})$, but explicit reparameterization is not applicable to the inverse-Wishart\footnote{Very recent work~\cite{Figurnov2018ImplicitR} may alleviate this issue and applying this technique is future work.}. It is also possible to employ a single variational distribution $q(\mathbf{z}|\mathbf{x})$ by recognizing that marginalization of the prior $p(\mathbf{z}|\bm{\Sigma})$ under an inverse-Wishart hyperprior leads to a multivariate Student's $t$-distribution. However, explicit reparameterization is also not applicable in this case and analytic marginalization may not be possible with arbitrary hyperprior specifications.

In the following, we describe approximate inference using a structured variational distribution $q(\mathbf{z},\bm{\Sigma}|\mathbf{x})=q(\mathbf{z}|\mathbf{x})q(\bm{\Sigma}|\mathbf{z})$. Unlike the VAE, we learn a full covariance matrix and  set the conditional $q(\mathbf{z}|\mathbf{x}) = \mathcal{N}(\mathbf{z}|\tilde{\bm{\mu}}, \tilde{\bm{\Sigma}} = \tilde{\mathbf{L}}\tilde{\mathbf{L}}^{\top})$ where $\tilde{\bm{\mu}}$ is the mean vector and $\tilde{\mathbf{L}}$ is the Cholesky factor of the covariance matrix, $\tilde{\bm{\Sigma}} = \tilde{\mathbf{L}}\tilde{\mathbf{L}}^{\top}$. Both $\tilde{\bm{\mu}}$ and $\tilde{\mathbf{L}}$ are obtained via a neural network $f_{\bm{\phi}}(\mathbf{x})$. Samples of $\mathbf{z}$ from $q(\mathbf{z}|\mathbf{x})$ are obtained via explicit reparameterization,
\begin{align}
\mathbf{z} = \tilde{\bm{\mu}} + \tilde{\mathbf{L}}\bm{\varepsilon}
\end{align}
where $\bm{\varepsilon} \sim \mathcal{N}(\bm{0}, \mathbf{I})$. Next, we describe how $q(\bm{\Sigma}|\mathbf{z})$ can be estimated efficiently. $\mathcal{L}_{\mathrm{ELBO}}$ in Eq. (\ref{eq:elbo}) can be written as
\begin{align}
\nonumber\mathcal{L}_{\mathrm{ELBO}} &= 
\mathbb{E}_{q(\mathbf{z}|\mathbf{x})q(\bm{\Sigma}|\mathbf{z})}\left[\log \frac{p(\mathbf{x}|\mathbf{z})p(\mathbf{z})}{q(\mathbf{z}|\mathbf{x})} + \log \frac{p(\bm{\Sigma}|\mathbf{z})}{q(\bm{\Sigma}|\mathbf{z})}\right]\\\nonumber
&=\mathbb{E}_{q(\mathbf{z}|\mathbf{x})}\left[\log \frac{p(\mathbf{x}|\mathbf{z})p(\mathbf{z})}{q(\mathbf{z}|\mathbf{x})}\right]\\
&\quad- \mathbb{E}_{q(\mathbf{z}|\mathbf{x})}\left[D_{\mathrm{KL}}({q(\bm{\Sigma}|\mathbf{z})}||{p(\bm{\Sigma}|\mathbf{z})})\right].
\label{eq:elboalternate}
\end{align}
The first term in Eq. (\ref{eq:elboalternate}) is independent of $\bm{\Sigma}$ and the second term is non-negative. As such, $\mathcal{L}_{\mathrm{ELBO}}$ is maximized when $q(\bm{\Sigma}|\mathbf{z})$ matches $p(\bm{\Sigma}|\mathbf{z})$. With this in mind,
\begin{align}
q(\bm{\Sigma}|\mathbf{z}) \approx p(\bm{\Sigma}|\mathbf{z}) = \frac{p(\mathbf{z}|\bm{\Sigma})p(\bm{\Sigma})}{\int_{\bm{\Sigma}'}p(\mathbf{z}|\bm{\Sigma}')p(\bm{\Sigma}')d\bm{\Sigma}'}
\label{eq:bayesthm}
\end{align}
For $p(\mathbf{z}|\bm{\Sigma}) = \mathcal{N}(\mathbf{z}|\bm{0},\bm{\Sigma})$, $p(\bm{\Sigma}) = \mathcal{W}_p^{-1}(\bm{\Sigma}|\bm{\Psi}, \nu)$, and a sample $ \mathbf{z}_{i} \sim p(\mathbf{z}|\bm{\Sigma})$, we can  exploit the fact that the inverse-Wishart is a conjugate prior for the multivariate normal. We marginalize $\bm{\Sigma}'$ from the denominator in Eq. (\ref{eq:bayesthm}) and obtain $p(\bm{\Sigma}|\mathbf{z}) = \mathcal{W}_p^{-1}(\bm{\Psi} + \mathbf{z}_{i}\mathbf{z}_{i}^\top, \nu + 1)$. Using this distribution for $q(\bm{\Sigma}|\mathbf{z})$, we now write the ELBO as
\begin{align}
\nonumber\mathcal{L}_{\mathrm{ELBO}} &= \mathbb{E}_{q(\mathbf{z}|\mathbf{x})}\left[\log p(\mathbf{x}|\mathbf{z})\right]\\
&\quad-\mathbb{E}_{p(\bm{\Sigma}|\mathbf{z})}\left[D_{\mathrm{KL}}(\mathcal{N}(\tilde{\bm{\mu}}, \tilde{\bm{\Sigma}})\|\mathcal{N}(\bm{0}, \bm{\Sigma}))\right]\\\nonumber
&\quad-\mathbb{E}_{q(\mathbf{z}|\mathbf{x})}\left[D_{\mathrm{KL}}(\mathcal{W}_p^{-1}(\bm{\Phi}, \lambda)\|\mathcal{W}_p^{-1}(\bm{\Psi}, \nu))\right]
\label{eq:elbofinal}
\end{align}
where $\bm{\Phi} = \bm{\Psi} + \mathbf{z}_{i}\mathbf{z}_{i}^\top$ and $\lambda = \nu + 1$.

All three terms in the lower bound above have closed-form expressions and can be computed in a straight-forward manner (please refer to the supplementary material\footnote{Supplementary material for this paper is available at \url{https://arxiv.org/abs/1809.04497}} for detailed expressions). The first term is the reconstruction error, similar to other VAE based models. The second term represents the distance from the prior and discourages the latent codes from being too far away from the zero mean prior (this enables sampling and ensures that CHyVAE remains a valid generative model). The third term is an additional penalty on the covariance matrix; to encourage disentanglement, the prior is set as the identity matrix $\mathbf{I}$. As previously mentioned, when $\nu$ is increased, independence in the latent dimensions is more enforced, leading to disentangled representations. 

\subsubsection{Sample Generation} Generally, we use Bartlett decomposition to obtain samples from the inverse-Wishart distribution (more details in the supplementary material). For models trained with large values of $\nu$, we found that directly sampling from $\mathcal{N}(\mathbf{0}, \mathbf{I})$ also generates good images.

\section{Related Work}
Early works that have demonstrated disentanglement in limited settings include \cite{Schmidhuber1992LearningFC,Glorot2011DomainAF,Desjardins2012DisentanglingFO}, and several prior research has addressed the problem of disentanglement in \emph{supervised} or \emph{semi-supervised} settings \cite{Kulkarni2015DeepCI,Kingma2014SemiSupervisedLW,Reed2014LearningTD,Siddharth2017LearningDR}. In this work, we focus on unsupervised learning of disentangled features. Unsupervised generative models such as \cite{VAE_Kingma2013AutoEncodingVB,Makhzani2015AdversarialA,Radford2015UnsupervisedRL} have been also shown to learn disentangled representations, although this was not the main motivation of these works. Recent work has also sought to disentangle factors of variation in sequential data in an unsupervised manner \cite{Denton2017UnsupervisedLO,Hsu2017UnsupervisedLO}.

\paragraph{VAE-based Models} Our work builds upon the VAE and is inspired by recent work on learning disentangled factors~\cite{BetaVAE_Matthey2016vaeLB,FactorVAE_Kim2018DisentanglingBF,Chen2018IsolatingSO}. In addition to these papers (covered in Sec. \ref{sec:background}), there has been further work in uncovering the principles behind disentanglement in VAEs. \citeauthor{Burgess2017UnderstandingDI} \shortcite{Burgess2017UnderstandingDI} argue from the information bottleneck principle that penalizing mutual information in $\beta$-VAE results in a compact and disentangled representation. Based on their analyses, \citeauthor{Kumar2017VariationalIO} \shortcite{Kumar2017VariationalIO} add an additional penalty to the VAE ELBO based on how much the covariance of $q(\mathbf{z})$ deviates from $\mathbf{I}$. In contrast to these previous work, we attempt to introduce disentanglement at the model specification stage through the covariance hyperprior.

\paragraph{InfoGAN} An alternative approach towards deep generative modeling is the Generative Adversarial Network (GAN)~\cite{GAN_NIPS2014_5423}. \citeauthor{InfoGAN_NIPS2016_6399} \shortcite{InfoGAN_NIPS2016_6399} have argued that maximizing mutual information between the observed sample and a subset of latent codes encourages disentanglement and capitalized on this idea to develop the InfoGAN. \citeauthor{FactorVAE_Kim2018DisentanglingBF} \shortcite{FactorVAE_Kim2018DisentanglingBF} evaluated the disentanglement performance of InfoWGAN-GP, a version of InfoGAN that uses WGAN \cite{WGAN_pmlr-v70-arjovsky17a} and gradient penalty \cite{GAN_GP_NIPS2017_7159}. 

\paragraph{Other priors} Previous work has explored different priors for the latent space in VAEs including mixture of Gaussians \cite{Jiang2016VariDeep}, Dirichlet process \cite{Nalisnick2016StickB}, Beta, Gamma, and von Mises \cite{Figurnov2018ImplicitR}. With advancements in reparameterization for discrete distributions, recent work  \cite{Esmaeili2018StructuredLR,Pineau2018InfoCatVAERL} have proposed adding different priors to different subsets of the latent code to separately model discrete and continuous factors of variation in the data. However, to the best of our knowledge, we are the first to apply hierarchical priors towards learning disentangled representations.

\section{Experiments}
In this section, we report on experiments comparing CHyVAE to two VAE-based unsupervised disentangling models: the $\beta$-VAE and state-of-the-art FactorVAE\footnote{We exclude comparisons with InfoGAN as the VAE-based models have been shown to obtain better disentanglement performance relative to InfoGAN in previous work~\cite{BetaVAE_Matthey2016vaeLB,FactorVAE_Kim2018DisentanglingBF}.}. Due to space constraints, we briefly describe the experimental setup and focus on the main findings; details are available in the supplementary material and our code base is available for download at \url{https://github.com/crslab/CHyVAE}.

\subsection{Experimental Setup}

\subsubsection{Model Implementation and Training}
To ease comparisons between the methods and prior work, we use the same network architecture across all the compared methods. Specifically, we follow the model in ~\cite{FactorVAE_Kim2018DisentanglingBF}: a convolutional neural network (CNN) for the encoder and a deconvolutional NN for the decoder. We normalize all datasets to $[0,1]$ and use sigmoid cross-entropy as the reconstruction loss function. For training, we use Adam optimizer \cite{Kingma2014AdamA} with a learning rate of $10^{-4}$. For the discriminator in FactorVAE, we use the parameters recommended by \citeauthor{FactorVAE_Kim2018DisentanglingBF} \shortcite{FactorVAE_Kim2018DisentanglingBF}. 

\subsubsection{Datasets}
Our experiments were conducted using five datasets, including four standard benchmarks:
\begin{enumerate}[{1)}]
	\item Datasets with \emph{known} generative factors: 
	\begin{enumerate}
		\item \textbf{2DShapes} (or \textit{dSprites}) \cite{dsprites17}: 737,280 binary $64 \times 64$ images of 2D shapes (heart, square, ellipse) with five ground truth factors of variation namely x-position, y-position, scale, orientation, and shape. All factors except shape are continuous.
		\item \textbf{CorrelatedEllipses}: 200,000 grayscale $64 \times 64$ images of ellipses with dependent ground truth factors x-position correlated with y-position and scale correlated with orientation. This dataset embodies the ``adversarial'' case where the factors may be correlated or the dataset was obtained with sampling bias (arguably common in many real-world data). Dataset construction details are in the supplementary material. 
	\end{enumerate}
	\item Datasets with unknown generative factors:  
	\begin{enumerate}
		\item \textbf{3DFaces} \cite{Paysan2009A3F}: 239,840 greyscale $64 \times 64$ images of 3D Faces.
		\item \textbf{3DChairs} \cite{Aubry2014Seeing3C}: 86,366 RGB $64 \times 64$ images of CAD chair models. 
		\item \textbf{CelebA} \cite{Liu2015DeepLF}: 202,599 RGB images  of celebrity faces center-cropped to dimensions $64 \times 64$.
	\end{enumerate}
\end{enumerate}

\begin{figure}
	\centering
	\includegraphics[width=0.70\columnwidth]{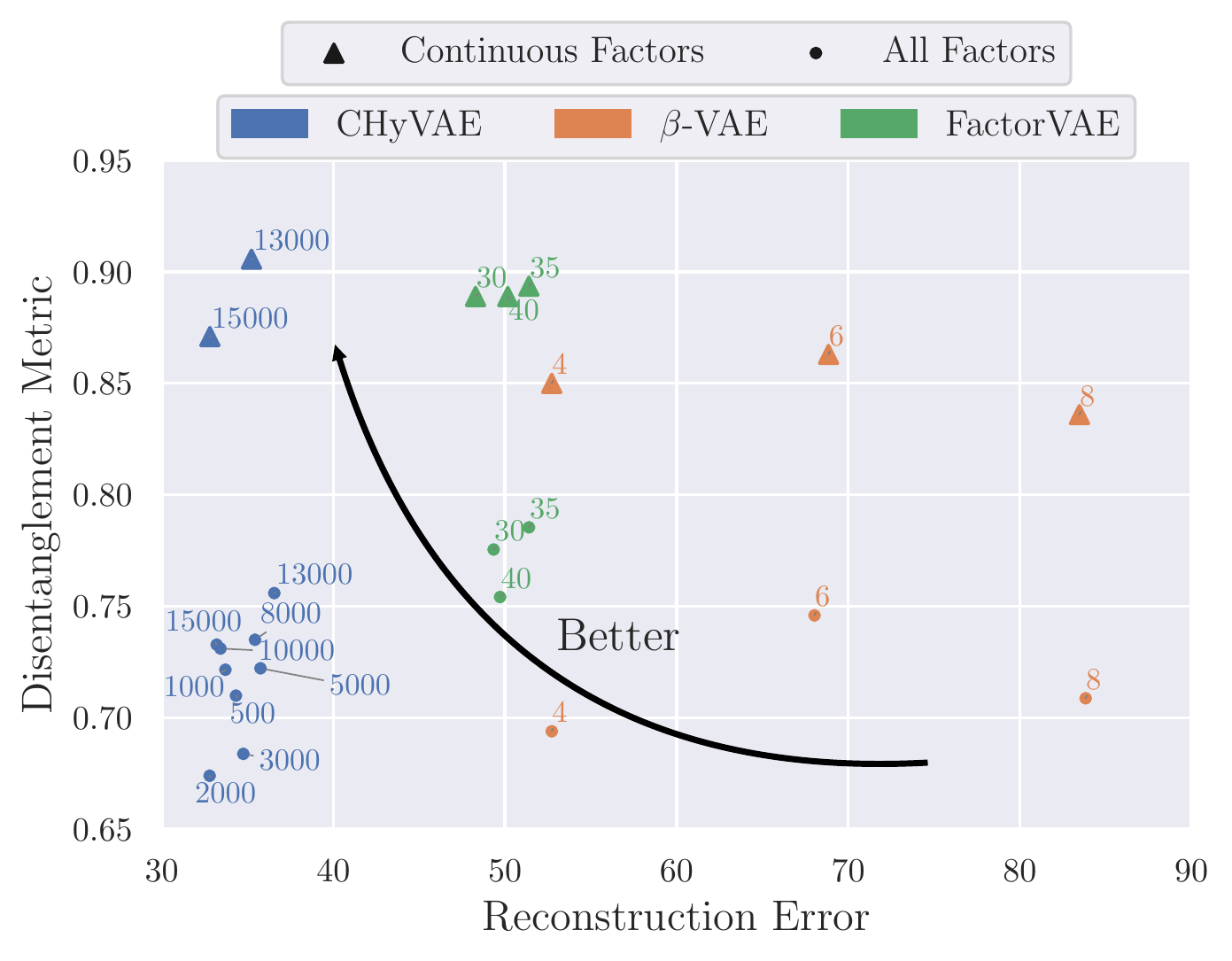}
	\caption{Disentanglement Metric plotted against Reconstruction error for CHyVAE, $\beta$-VAE, and FactorVAE on 2DShapes dataset averaged over 10 random restarts.}
	\label{fig:2d_shapes_plots}
\end{figure}
\begin{figure}
	\centering
	\includegraphics[width=0.70\columnwidth]{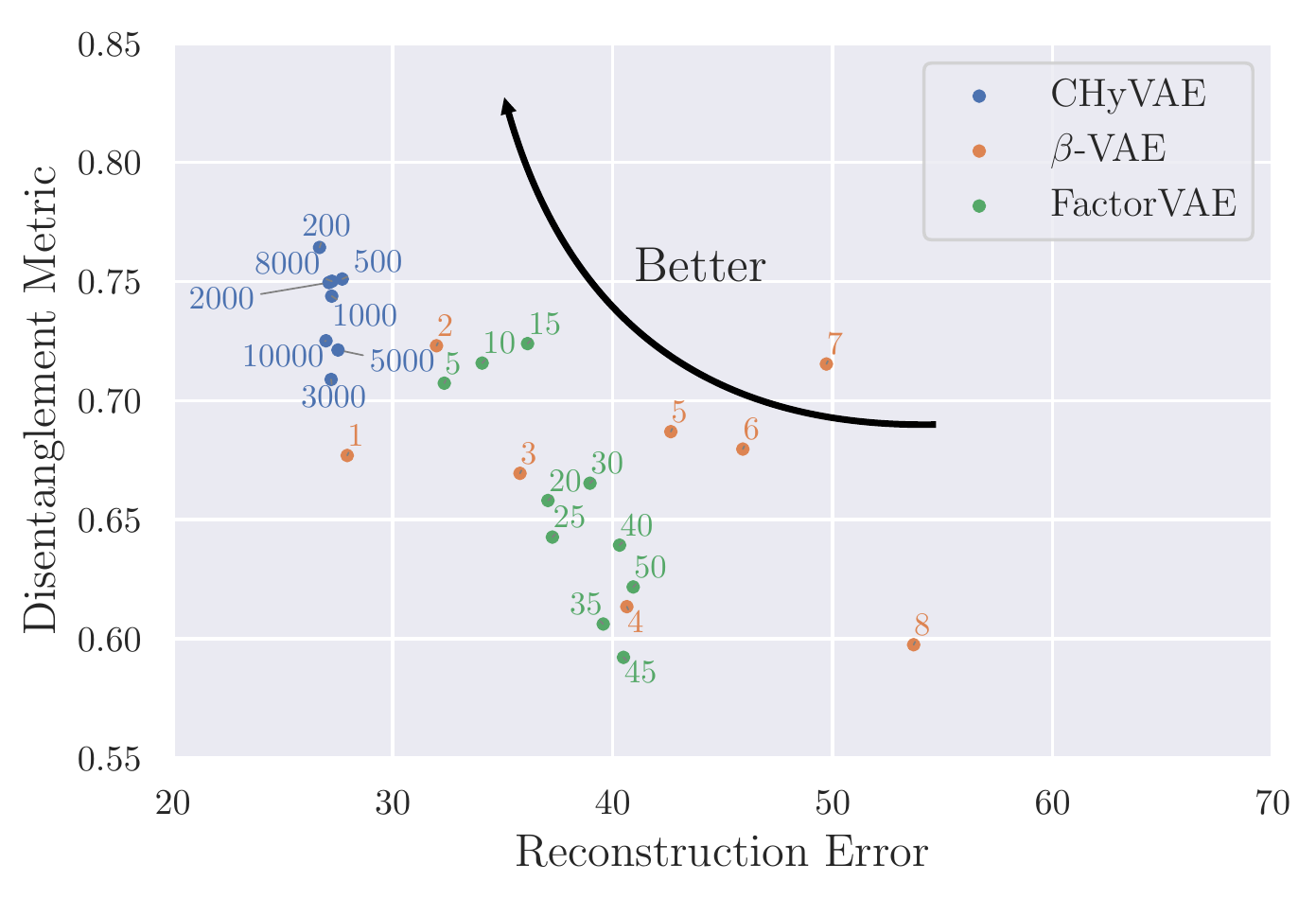}
	\caption{Disentanglement Metric plotted against Reconstruction error for CHyVAE, $\beta$-VAE, and FactorVAE on CorrelatedEllipses dataset averaged over 10 random restarts.}
	\label{fig:correll_avg}
\end{figure}

\begin{figure}
	\centering
	\includegraphics[width=0.80\columnwidth]{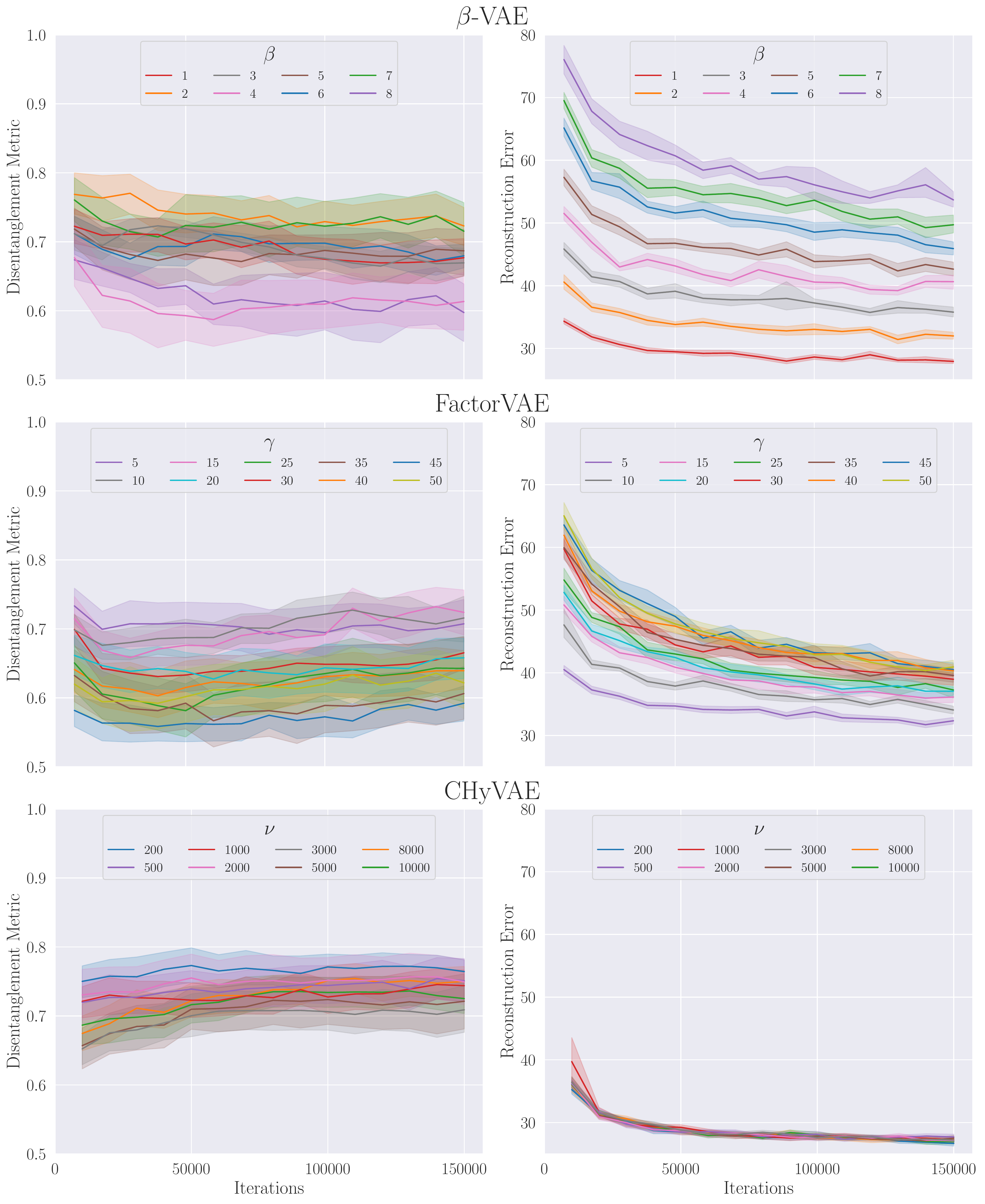}
	\caption{Disentanglement metric scores (left) and reconstruction error (right) with iterations for $\beta$-VAE, FactorVAE, and CHyVAE on CorrelatedEllipses dataset averaged over 10 random restarts.}
	\label{fig:corr_ellipses_plots}
\end{figure}

\subsubsection{Disentanglement Metric and Latent Traversals}
Suitable evaluation criteria for disentanglement remains an area of active research. Several metrics have been recently proposed based on linear mappings from latent codes to generative factors \cite{BetaVAE_Matthey2016vaeLB,eastwood2018a} and mutual information \cite{Chen2018IsolatingSO}. 

We use the metric proposed by \citeauthor{FactorVAE_Kim2018DisentanglingBF} \shortcite{FactorVAE_Kim2018DisentanglingBF} for evaluating the models primarily because of its interpretability and computational efficiency. The metric uses a majority vote classifier matrix $\mathbf{V}^{p \times K}$ that maps each latent dimension to only one ground truth factor where $p$ is the dimension of the latent code and $K$ is the number of ground truth factors. Each element $\mathbf{V}_{ij}$ for $i \in \{1 \dots p\}, j \in \{1 \dots K\}$ is a count of the number of batches with a fixed factor $j$ that have minimum variance in the dimension $i$ of the latent code. Using the vote matrix in the metric each latent dimension can be mapped to a ground truth factor and the dimensions can be annotated. Note that quantitative evaluation can only be performed on datasets with \emph{known} factors of variation.

When the factors of variation are \emph{unknown}, it is common to examine latent traversals. These traversals are obtained by fixing all latent dimensions and varying only one. Inspection of latent traversals tells us little about the robustness of a model but is currently the only available method of comparing disentanglement performance on datasets with unknown factors of variation.

\subsection{Quantitative Evaluation}

Our main results on the 2DShapes dataset are summarized in Figure \ref{fig:2d_shapes_plots}; it shows the disentanglement metric plotted against the reconstruction error (scores averaged over 10 random restarts) for varying values of $\beta$, $\gamma$, and $\nu$\footnote{For $\beta$-VAE and FactorVAE, we show results using the best performing hyperparameter values reported in \citeauthor{FactorVAE_Kim2018DisentanglingBF} \shortcite{FactorVAE_Kim2018DisentanglingBF}.}. Better performing methods  fall on the top left of the graph (high disentanglement with low reconstruction error). 

Figure \ref{fig:2d_shapes_plots} clearly shows that CHyVAE outperforms both $\beta$-VAE and FactorVAE on the continuous factors, achieving far better reconstruction error at similar---if not, slightly better---disentangling performance. Higher values of $\nu$ tend to produce better disentanglement, while preserving reconstruction capability. When the discrete factor (shape) is included, the FactorVAE achieves slightly higher disentanglement on average, but the best performing models are comparable---$0.909$ ($\gamma=35$) and $0.905$ ($\nu=13000$) for FactorVAE and CHyVAE respectively. However, the latent traversals show that all the models struggle with the discrete factor (in supplementary material). For the CHyVAE, it is unsurprising that the disentanglement would be poorer with discrete latent factors: CHyVAE enforces a hierarchical prior on a \emph{continuous} latent space. Discrete factors can handled in a principled manner within our framework by using a suitable prior~\cite{Esmaeili2018StructuredLR,Pineau2018InfoCatVAERL} with associated hyperprior.

\begin{figure}[t]
	\centering
	\subfloat{\includegraphics[width=0.80\columnwidth]{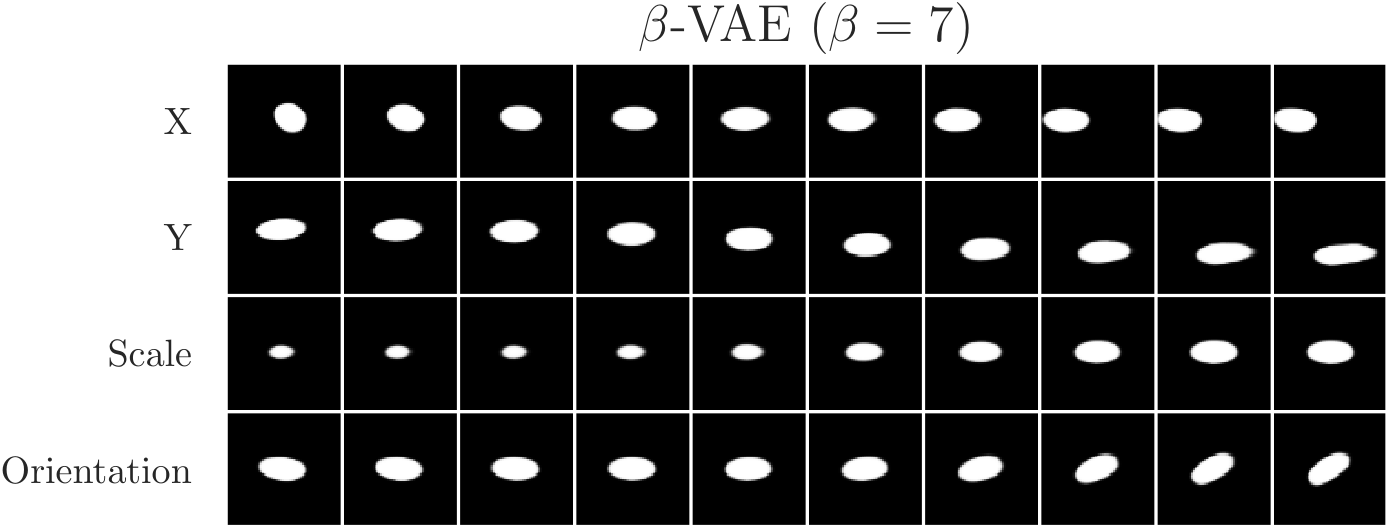}}\\
	\subfloat{\includegraphics[width=0.80\columnwidth]{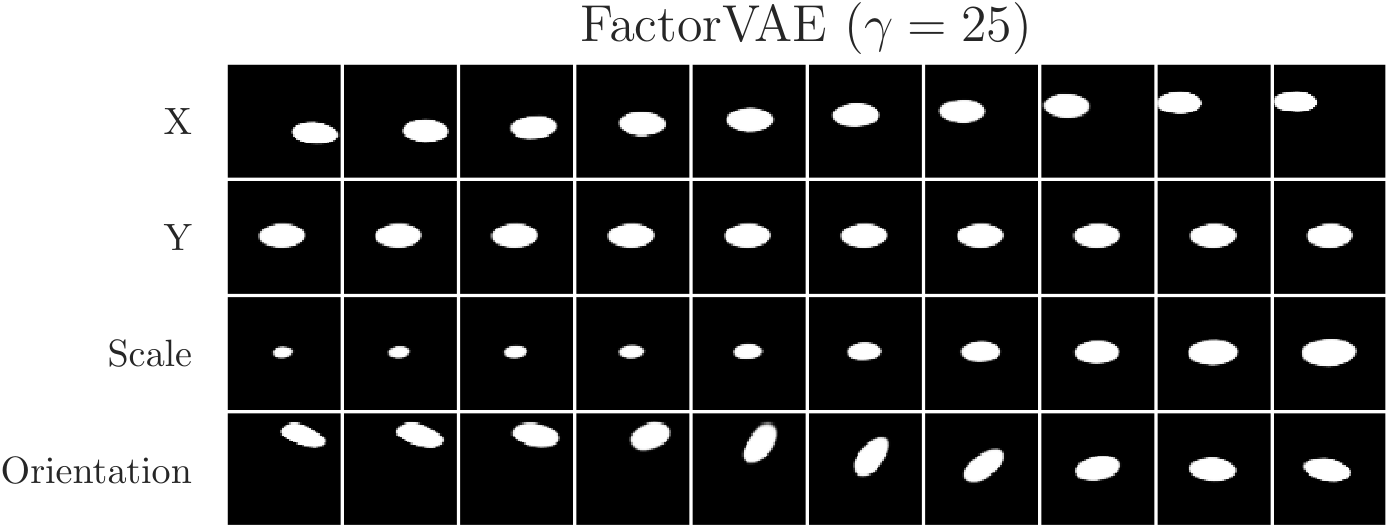}}\\
	\subfloat{\includegraphics[width=0.80\columnwidth]{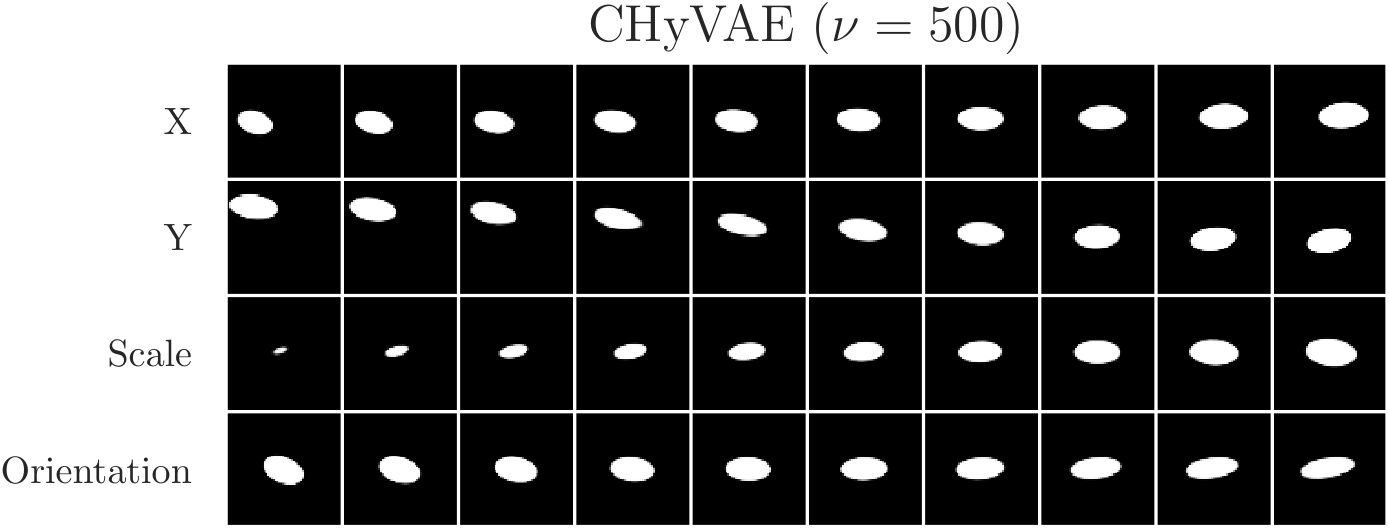}}
	\caption{Traversals across latent dimensions on CorrelatedEllipses dataset annotated with the factor of variation assigned by the majority vote matrix for best performing $\beta$-VAE, FactorVAE, and CHyVAE.}
	\label{fig:correll_chyvae_inter}
\end{figure}

Results for CorrelatedEllipses are summarized in Fig. \ref{fig:correll_avg}. CHyVAE starkly outperforms $\beta$-VAE and FactorVAE both in terms of the metric and the reconstruction error across the different parameters. We posit that this was due to the extra flexibility afforded by the prior and hyperprior and learning a full covariance matrix; lower values of $\nu$, which allow more deviation from the identity covariance, achieve better disentanglement and reconstruction.  Figure \ref{fig:corr_ellipses_plots} shows the the disentanglement metric and the reconstruction error as training progressed for different parameter values; compared to FactorVAE and $\beta$-VAE, CHyVAE achieves lower reconstruction error and better disentanglement. Figure \ref{fig:correll_chyvae_inter} shows the latent traversals for best performing models on the disentanglement metric. Interestingly, both $\beta$-VAE and CHyVAE learn a slightly-entangled representation for the y-position but the FactorVAE fails to capture this factor.

\begin{figure*}
	\centering
	\includegraphics[width=0.95\textwidth]{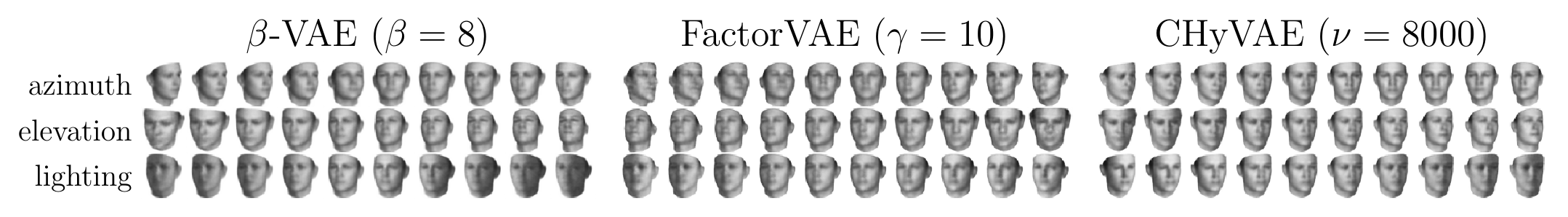}
	\caption{Traversals across latent dimensions for $\beta$-VAE, FactorVAE, and CHyVAE on 3DFaces dataset annotated with the factor of variation.}
	\label{fig:3dfaces_chyvae_inter}
\end{figure*}

\subsection{Qualitative Evaluation}

\begin{figure}
	\centering
	\subfloat{\includegraphics[width=0.84\columnwidth]{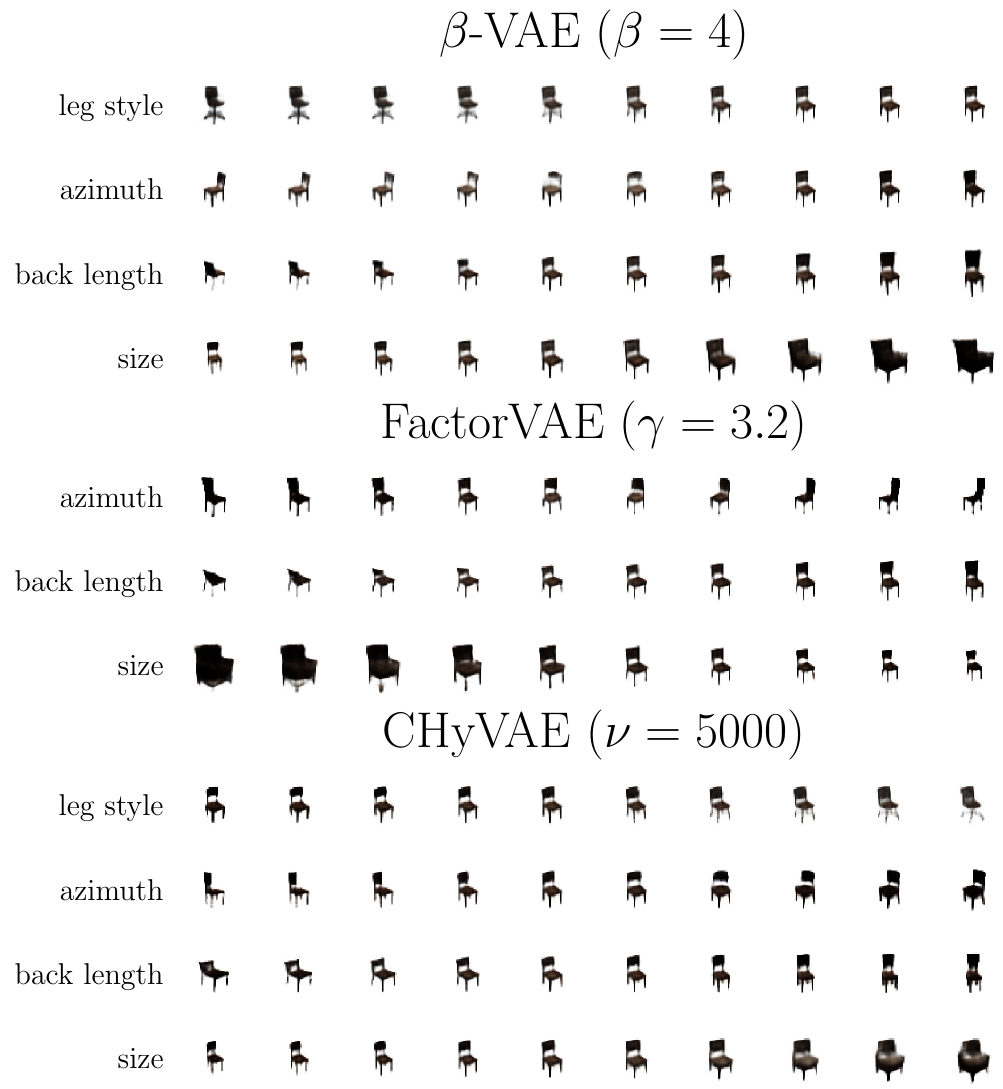}}\\
	\caption{Traversals across latent dimensions for $\beta$-VAE, FactorVAE, and CHyVAE on 3DChairs dataset annotated with the factor of variation.}
	\label{fig:3dchairs_chyvae_inter}
\end{figure}

\begin{figure}
	\centering
	\includegraphics[width=0.70\columnwidth]{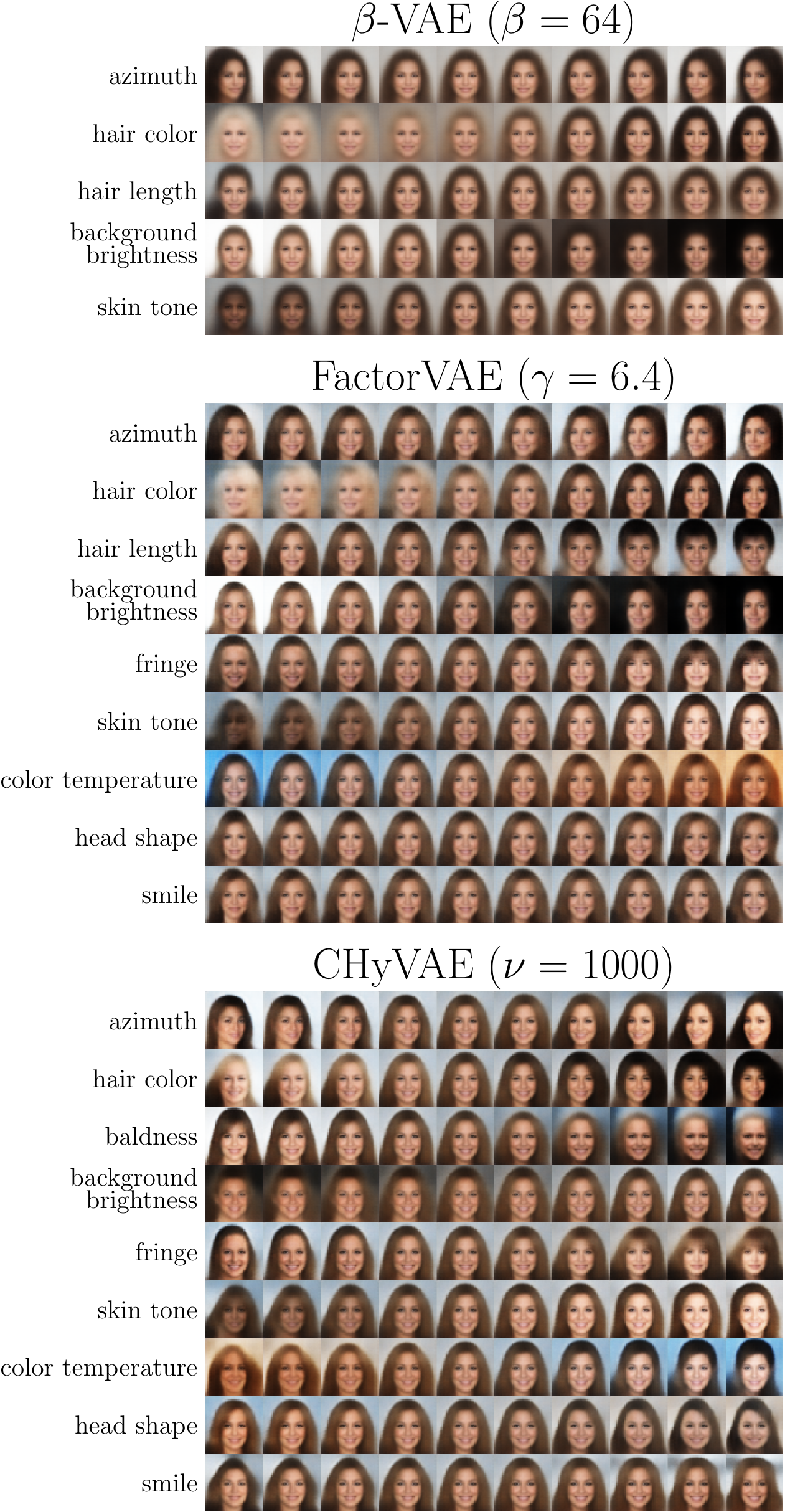}
	\caption{Traversals across latent dimensions for $\beta$-VAE, FactorVAE, and CHyVAE on CelebA dataset annotated with the most prominently varying factor of variation.}
	\label{fig:celeba_chyvae_inter}
\end{figure}

\begin{figure}
	\centering
	\includegraphics[width=0.60\columnwidth]{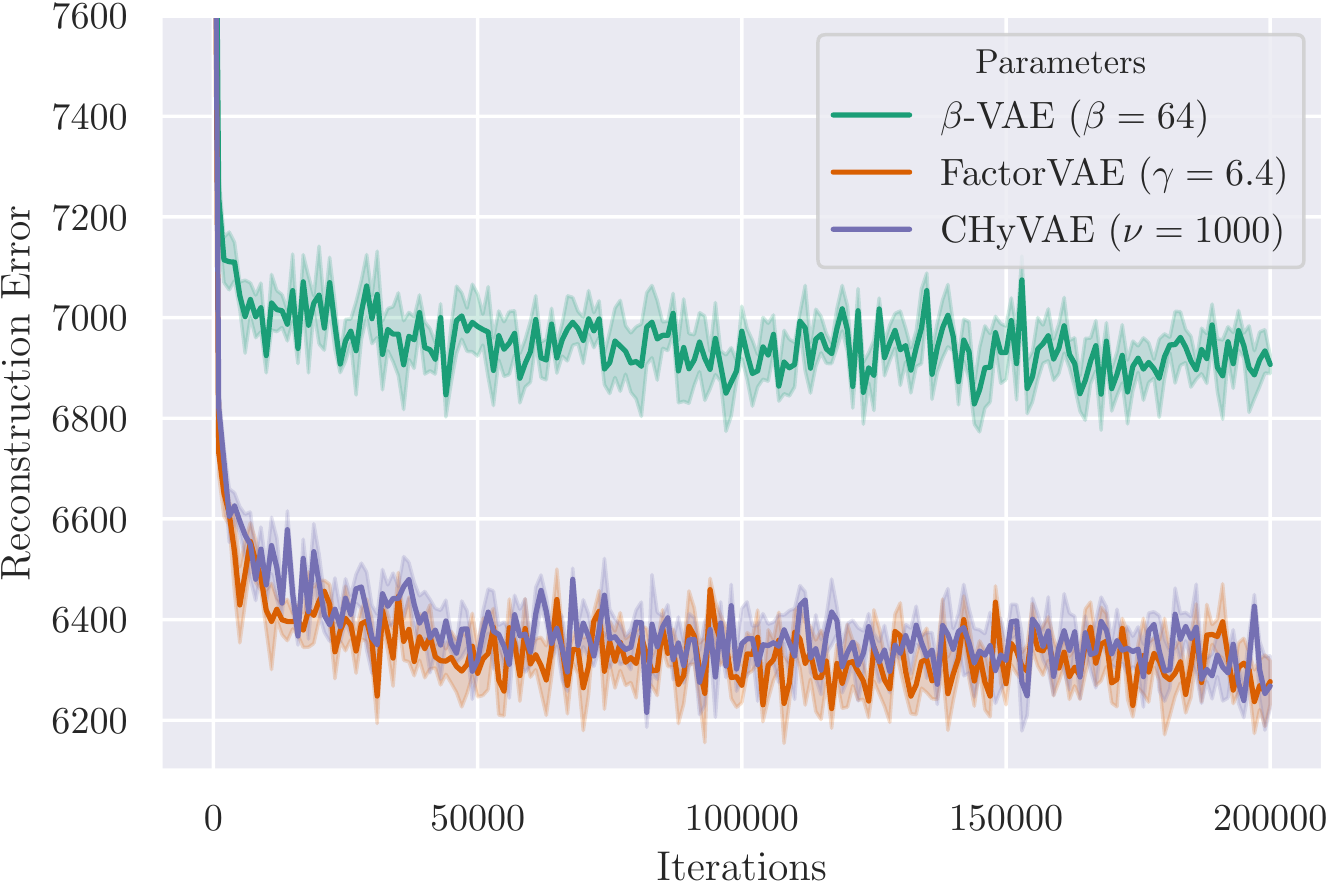}
	\caption{Reconstruction error vs iterations for $\beta$-VAE, FactorVAE, and CHyVAE on CelebA averaged over 5 random restarts.}
	\label{fig:celeba_recons}
\end{figure}

In absence of a metric for comparison of the disentangling performance of different models on datasets with unknown generative factors, the only evaluation method available is inspecting latent traversals. Figs. \ref{fig:3dfaces_chyvae_inter} and \ref{fig:celeba_chyvae_inter} show that CHyVAE is able to learn semantically reasonable factors of variation for 3DFaces and CelebA. For 3DChairs (Fig. \ref{fig:3dchairs_chyvae_inter}) CHyVAE is able to learn the \textit{leg-style} factor which is missed by FactorVAE but learnt by the $\beta$-VAE. In terms of reconstruction, CHyVAE achieves superior performance relative to $\beta$-VAE and comparable to FactorVAE; see Fig. \ref{fig:celeba_recons} and refer the supplementary material for plots for 3DFaces and 3DChairs. 

\section{Conclusion}
State-of-the-art methods for learning disentangled representations in VAEs have focussed primarily on manipulating the ELBO. In contrast, we pursued an alternative principled approach by placing a hyperprior on the covariance matrix of the VAE prior. The inverse-Wishart used in our study exposes its degrees-of-freedom parameter which can be tuned to control the informativeness of a desired independent covariance $\mathbf{I}$ and thus, encourage disentanglement.

Extensive experiments on a variety of datasets show that our model, CHyVAE, outperforms the $\beta$-VAE and is comparable to the FactorVAE in terms of disentanglement, while achieving better reconstruction. Our experimental results with a new dataset also demonstrate that encouraging factorial codes may not learn suitable disentangled representations when correlations are present; instead, a more flexible model such as CHyVAE may disentangle better.

While we have focussed on the inverse-Wishart hyperprior in this work, our key idea of using a hierarchical model can be extended to alternative distributions. As future work, we plan to examine the effects of different hyperpriors, and extend the approach towards learning  disentangled representations with both discrete and continuous latent variables. We also plan to explore technical improvements, e.g., the structured factorization explored in this work is an improvement over standard mean-field, but it may lose information due to the choice of simplified distributions and the optimality assumption underlying the approximation of $q(\bm{\Sigma}|\mathbf{z})$. 

\fontsize{9.0pt}{10.0pt}\selectfont 
\bibliographystyle{aaai}
\bibliography{aaai.bib}
\appendix
\setcounter{secnumdepth}{2}
\section{Appendix}
\begin{table*}
	\centering
	\subfloat[Architecture for 2DShapes and CorrelatedEllipses]{\begin{tabular}{|l|l||l|l|}
			\hline
			\multicolumn{2}{|c||}{\textbf{Encoder}}	& \multicolumn{2}{|c|}{\textbf{Decoder}} \\
			\hline
			Layer & Activation & Layer & Activation\\
			\hline
			Input $64\times 64 \times 1$ & --    & Input 10 & --\\
			\hline
			Conv $32\times 4 \times 4$ (stride 2)	& $\mathrm{ReLU}$	& Dense 128	& $\mathrm{ReLU}$	 \\
			\hline
			Conv $32\times 4 \times 4$ (stride 2)	& $\mathrm{ReLU}$	& Dense 1024 & $\mathrm{ReLU}$	 \\
			\hline
			Conv $64\times 4 \times 4$ (stride 2)	& $\mathrm{ReLU}$	& UpConv $64\times 4 \times 4$ (stride 2) & $\mathrm{ReLU}$ \\
			\hline
			Conv $64\times 4 \times 4$ (stride 2)	& $\mathrm{ReLU}$	& UpConv $32\times 4 \times 4$ (stride 2) & $\mathrm{ReLU}$ \\
			\hline
			Dense 128	& $\mathrm{ReLU}$	& UpConv $32\times 4 \times 4$ (stride 2) & $\mathrm{ReLU}$ \\
			\hline
			$\bm{\mu}$ = Dense 10	& --	&   & \multirow{3}{*}{Sigmoid}\\
			$\mathbf{A}$ = Dense $10\times 10$ (CHyVAE)	& --	& \multicolumn{1}{c|}{UpConv $1\times 4 \times 4$ (stride 2)} &	 \\
			$\bm{\sigma}$ = Dense 10 ($\beta$-VAE, FactorVAE)	& $\mathrm{softplus}$	&  &	 \\
			\hline
			
	\end{tabular}}
	
	\subfloat[Architecture for 3DFaces, 3DChairs, and CelebA]{\begin{tabular}{|l|l||l|l|}
			\hline
			\multicolumn{2}{|c||}{\textbf{Encoder}}	& \multicolumn{2}{|c|}{\textbf{Decoder}} \\
			\hline
			Layer & Activation & Layer & Activation\\
			\hline
			Input $64\times 64 \times 1$ or $64\times 64 \times 3$& --    & Input 32 & --\\
			\hline
			Conv $32\times 4 \times 4$ (stride 2)	& $\mathrm{ReLU}$	& Dense 256	& $\mathrm{ReLU}$	 \\
			\hline
			Conv $32\times 4 \times 4$ (stride 2)	& $\mathrm{ReLU}$	& Dense 1024 & $\mathrm{ReLU}$	 \\
			\hline
			Conv $64\times 4 \times 4$ (stride 2)	& $\mathrm{ReLU}$	& UpConv $64\times 4 \times 4$ (stride 2) & $\mathrm{ReLU}$ \\
			\hline
			Conv $64\times 4 \times 4$ (stride 2)	& $\mathrm{ReLU}$	& UpConv $32\times 4 \times 4$ (stride 2) & $\mathrm{ReLU}$ \\
			\hline
			Dense 256	& $\mathrm{ReLU}$	& UpConv $32\times 4 \times 4$ (stride 2) & $\mathrm{ReLU}$ \\
			\hline
			$\bm{\mu}$ = Dense 32	& --	& UpConv $1\times 4 \times 4$ (stride 2) & \multirow{3}{*}{Sigmoid}\\
			$\mathbf{A}$ = Dense $32\times 32$ (CHyVAE)	& --	& \multicolumn{1}{c|}{or} &	 \\
			$\bm{\sigma}$ = Dense 32 ($\beta$-VAE, FactorVAE)	& $\mathrm{softplus}$	& UpConv $3\times 4 \times 4$ (stride 2) &	 \\
			\hline
			
	\end{tabular}}
	\caption{Encoder and decoder network architecture}
	\label{tab:archi}
\end{table*}
\subsection{Network and Hyperparameter Details}
\label{sec:networkdetails}
We use the same network architecture as \cite{FactorVAE_Kim2018DisentanglingBF} for 2DShapes and CorrelatedEllipses for a fair comparison on the disentanglement metric. For 3DFaces, 3DChairs, and CelebA we use the same encoder and decoder architecture as \cite{FactorVAE_Kim2018DisentanglingBF} with a 32 dimensional latent space. The discriminator network in FactorVAE is constructed and trained as suggested by \citeauthor{FactorVAE_Kim2018DisentanglingBF}: a 6 layer MLP with 1000 units in each layer and leakyReLU activation. The encoder and decoder network architecture is summarized in table \ref{tab:archi}. We use Adam optimizer \cite{Kingma2014AdamA} with a learning rate of $10^{-4}$, $\beta_1=0.9$, and $\beta_2=0.999$ for training the VAEs. We use a batch size of $50$ and train for $150000$ (2DShapes, CorrelatedEllipses) or $200000$ (3DFaces, 3DChairs, CelebA) steps.
\paragraph{Learning the Cholesky factor} We now explain how to learn the Cholesky factor $\tilde{\mathbf{L}}$ in Eq. (9) in the paper. Learn a $p \times p$ matrix $\mathbf{A}$ using the encoder (see table \ref{tab:archi}); convert $\mathbf{A}$ to a lower triangular matrix $\mathbf{B}$ by zeroing the upper triangular portion; apply $\mathrm{softplus}$ to the diagonal elements of the lower triangular matrix $\mathbf{B}$ to get $\tilde{\mathbf{L}}$; $\tilde{\bm{\Sigma}}$ can now be obtained using $\tilde{\bm{\Sigma}}=\tilde{\mathbf{L}} \tilde{\mathbf{L}}^\top$.

\paragraph{Numerical Issues} While theoretically positive-definite, $\tilde{\bm{\Sigma}}$ computed as described above might not be numerically positive-definite, especially in \texttt{float32} precision; therefore, we use a numerically stable version $\tilde{\bm{\Sigma}}=\tilde{\mathbf{L}}\tilde{\mathbf{L}}^\top + 10^{-4}\mathbf{I}$. Another numerical instability arises during the computation of log-determinant of a matrix. For a matrix $\bm{\Omega}$, log-determinant can be computed as follows. Compute the Cholesky factor $\mathbf{K}$ of $\bm{\Omega}$; compute $\log |\bm{\Omega}| = 2 \sum_i \log \mathbf{K}_{ii}$, where $\mathbf{K}_{ii}$ is the \textit{i}-th diagonal element.
\paragraph{Hyperparameters for CHyVAE} The value of degrees-of-freedom $\nu$ for the IW distribution was tuned from the set $\{35, 50, 100, 200, 500, 1000, 2000, 3000, 5000, 8000,$ $10000, 13000, 15000\}$ for different datasets. For a given value of $\nu$, the scale matrix $\bm{\Psi}$ can be set as the following
\begin{align*}
\bm{\Psi} = (\nu-p-1)\bm{\Sigma}_0
\end{align*}
where $p$ is the dimension of the latent space and $\bm{\Sigma}_0$ is the desired covariance matrix which is set to $\mathbf{I}$ (identity).
\subsection{Metric Details}
\label{sec:metricdetails}
\cite{FactorVAE_Kim2018DisentanglingBF}'s disentanglement metric can be computed as follows.
\begin{enumerate}
	\item Choose a factor $k \in \{1\dots K\}$ where $K$ is the number of factors.
	\item Fix this factor $k$ and generate a batch of size $L$ with all other factors randomly varying.
	\item Obtain the latent codes from the model for the batch.
	\item Normalize each dimension by its empirical standard deviation taken over a batch of $M$ images.
	\item Take the empirical variance in each dimension.
	\item Denote the index of the dimension with minimum variance as $d$. The pair $(d,k)$ forms one data-point. 
	\item Repeat steps 1--6 to construct a batch $S = \{(d_b, k_b)\}_{b=1}^B$ of $B$ such pairs.
	\item Construct a vote matrix $\mathbf{V}^{p\times K}$ using $S$ where $p$ is the size of latent code and $K$ is the number of factors. Each element $\mathbf{V}_{ij}$ for $i \in \{1 \dots p\}, j \in \{1 \dots K\}$ is a count of the number of batches with a fixed factor $j$ that have minimum variance in the dimension $i$ of the latent code. Formally, \[\mathbf{V}_{ij} = \sum_{b=1}^B \mathbbm{1}(d_b=i,k_b=j)\] 
	\item Once the vote matrix is constructed, sample another test batch $T = \{(d_n, k_n)\}_{n=1}^N$ of size $N$ by repeating steps 1--6. The value of metric is then just the average classification accuracy
	\[
	\frac{1}{N}\sum_{n=1}^N \mathbbm{1}(k_n = \mathrm{arg max}_{j}\mathbf{V}_{d_n, j})
	\]
	
\end{enumerate}
For our experiments we set $L=200, M=5000, B=800$, and $N=800$.
\begin{figure}
	\centering
	\includegraphics[width=0.8\columnwidth]{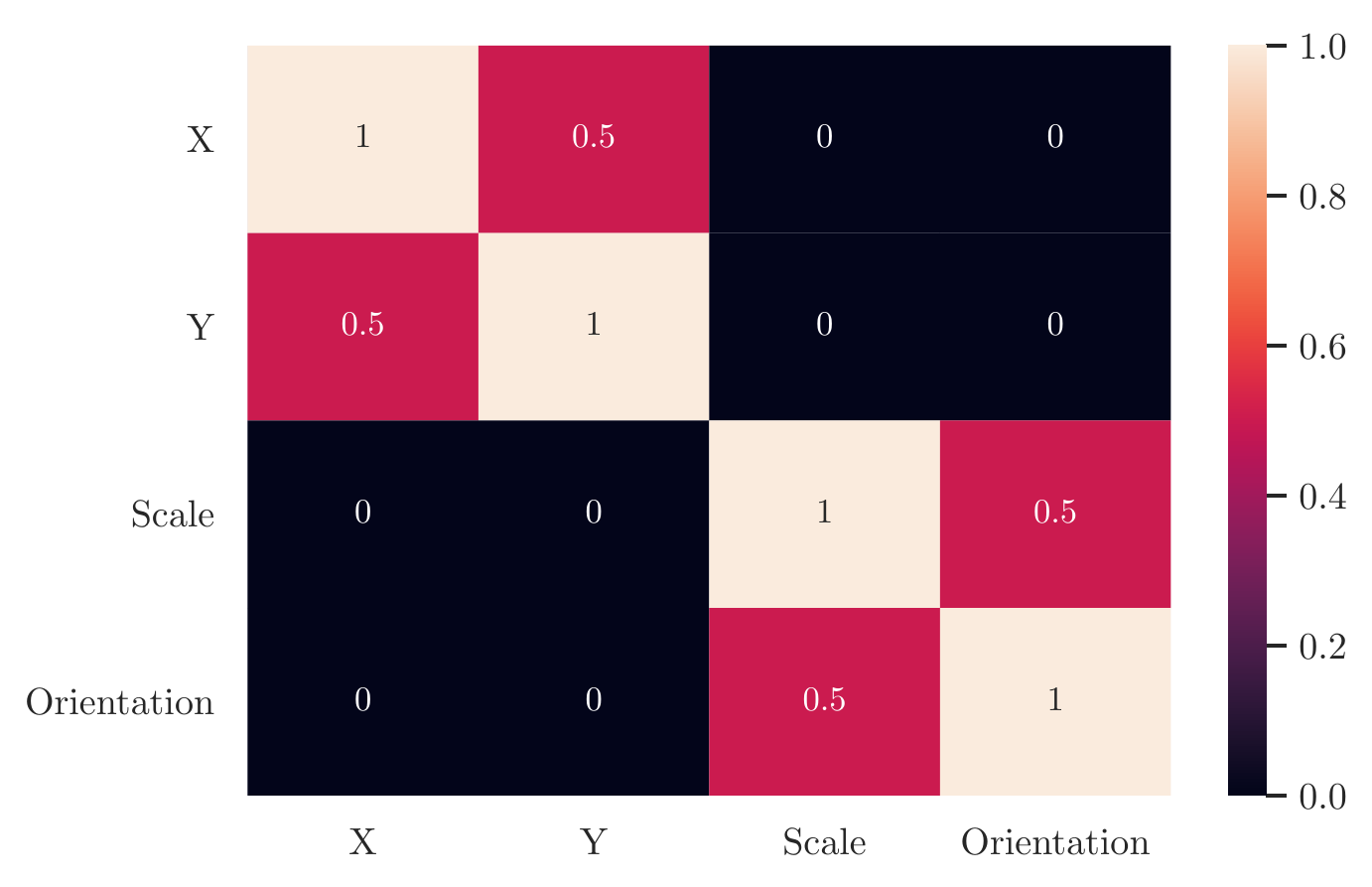}
	\caption{Block diagonal covariance matrix for Correlated Ellipses}
	\label{fig:corrcov}
\end{figure}
\subsection{Correlated Ellipses}
\label{sec:correlatedell}
We now explain how to generate a random sample for CorrelatedEllipses. 
\begin{itemize}
	\item Sample a vector $\mathbf{y}$ from a multivariate normal distribution with zero mean and block diagonal covariance as shown in figure \ref{fig:corrcov}.
	\item Clip $\mathbf{y}$ between $[-1,1]$ and then scale in $[0,1]$.
	\item Bin each element of $\mathbf{y}$ based on the possible values for each factor
	\begin{itemize}
		\item x-position: 32 values linearly spaced in $[0,1]$
		\item y-position: 32 values linearly spaced in $[0,1]$
		\item scale: 6 values linearly spaced in $[0.5,1]$
		\item orientation: 40 values linearly spaced in $[0,2\pi]$
	\end{itemize}
	Our choice of number of values for each factor is based on \cite{dsprites17}.
	\item Use the binned values to render images of ellipses using a suitable 2D rendering engine (such as OpenCV).
\end{itemize}

\begin{figure}
	\centering
	\subfloat{\includegraphics[width=0.6\columnwidth]{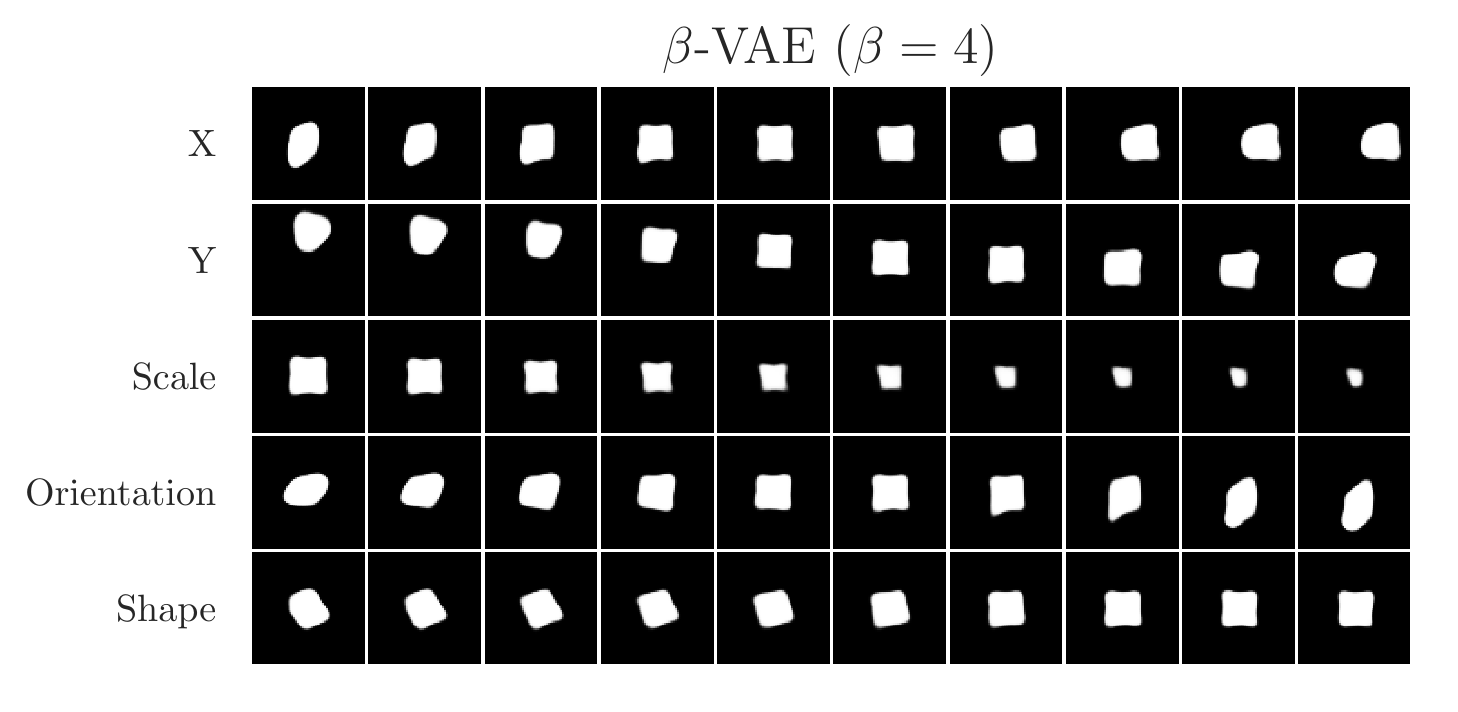}}\\
	\subfloat{\includegraphics[width=0.6\columnwidth]{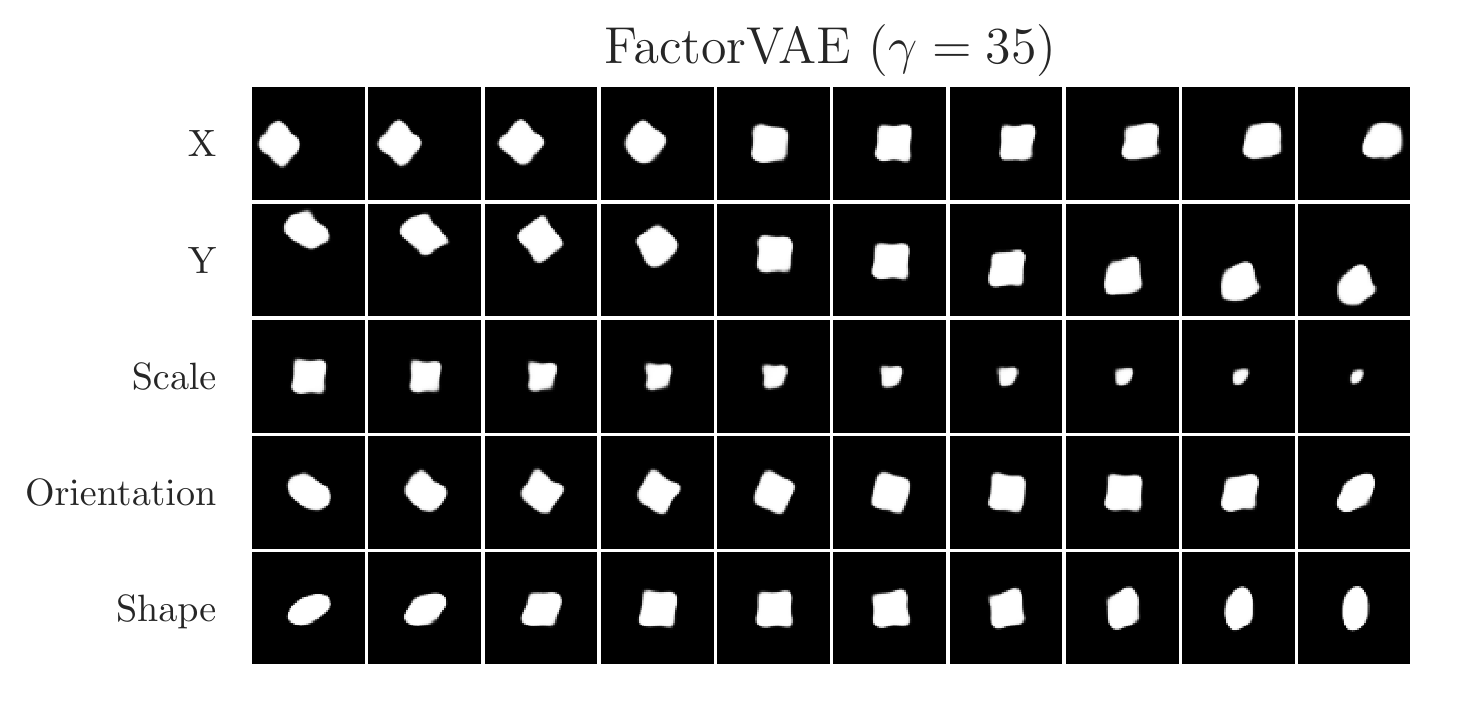}}\\
	\subfloat{\includegraphics[width=0.6\columnwidth]{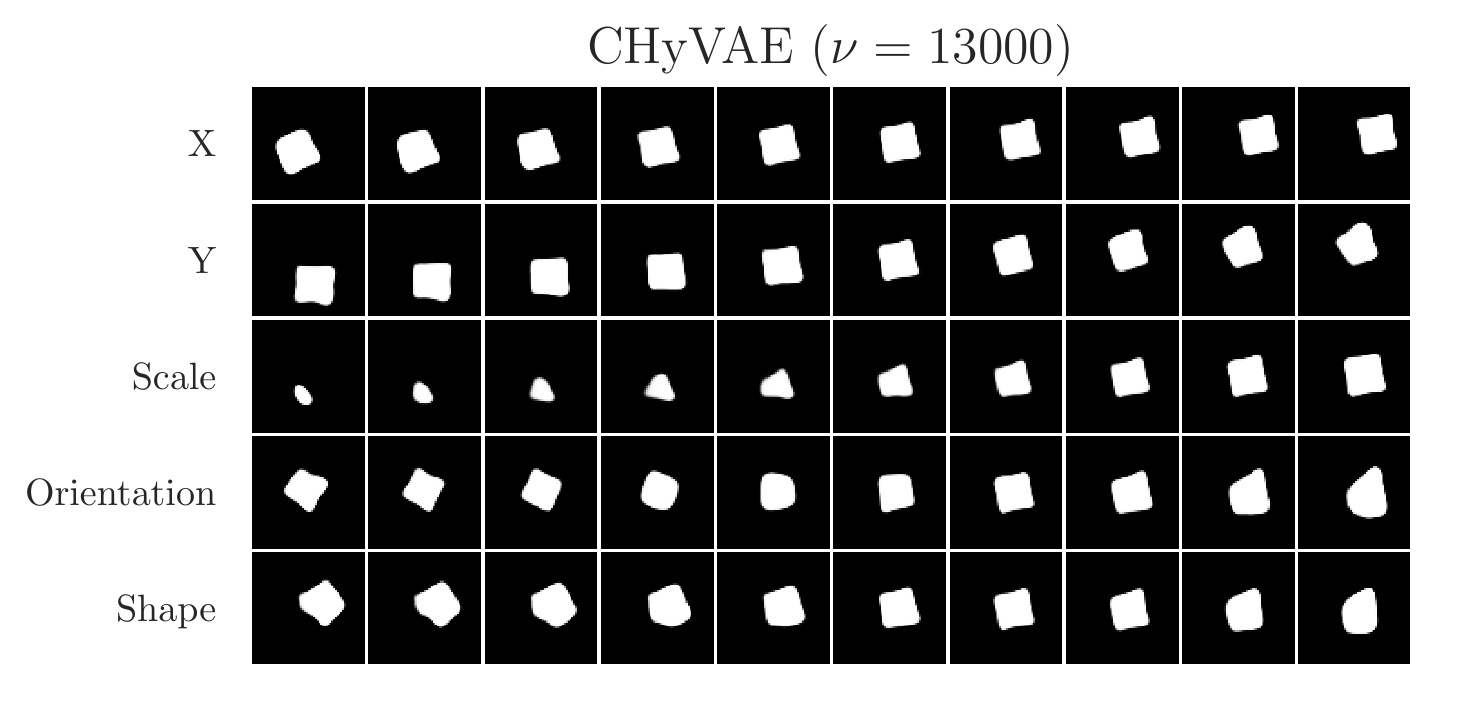}}
	\caption{Traversals across latent dimensions on 2DShapes dataset annotated with the factor of variation assigned by the majority vote matrix for $\beta$-VAE, FactorVAE, and CHyVAE with the best performance on the metric}
	\label{fig:2dshapes_chyvae_inter}
\end{figure}

\subsection{Sampling from Inverse-Wishart Distribution}
\label{sec:samplingIW}
Random samples from the an inverse-Wishart (IW) distribution can be generated using Bartlett decomposition utilizing the property that for $\mathbf{X} \sim \mathcal{W}(\bm{\Psi}^{-1}, \nu)$, $\mathbf{X}^{-1}$ is a sample from $\mathcal{W}^{-1}(\bm{\Psi}, \nu)$ where $\mathcal{W}$ and $\mathcal{W}^{-1}$ are the Wishart and inverse-Wishart distributions respectively. 

We now describe how to generate a sample from a Wishart distribution $\mathcal{W}(\bm{\Psi}^{-1}, \nu)$ using Bartlett decomposition \cite{wiki:wishart}.
\begin{itemize}
	\item Construct a matrix $\mathbf{B}$ as follows.
	\begin{align*}
	\mathbf{B} = \begin{bmatrix}
	c_{11} & 0 & 0 & \dots & 0\\
	n_{21} & c_{22} & 0 & \dots & 0\\
	n_{31} & n_{32} & c_{33} & \dots & 0\\
	\vdots & \vdots & \vdots & \ddots & \vdots\\
	n_{p1} & n_{p2} & c_{p3} & \dots & c_{pp}\\
	\end{bmatrix}
	\end{align*}
	where $n_{ij} \sim \mathcal{N}(0,1)$ and $c_{ii}^2 \sim \chi^2(\nu-i+1)$.
	\item Compute the Cholesky factor $\mathbf{V}$ of $\bm{\Psi}^{-1}$.
	\item $\mathbf{X} = \mathbf{V}\mathbf{B}\mathbf{B}^\top\mathbf{V}^\top$ is a sample from $\mathcal{W}(\bm{\Psi}^{-1}, \nu)$.
\end{itemize}
\begin{figure*}
	\centering
	\subfloat{\includegraphics[width=0.32\textwidth]{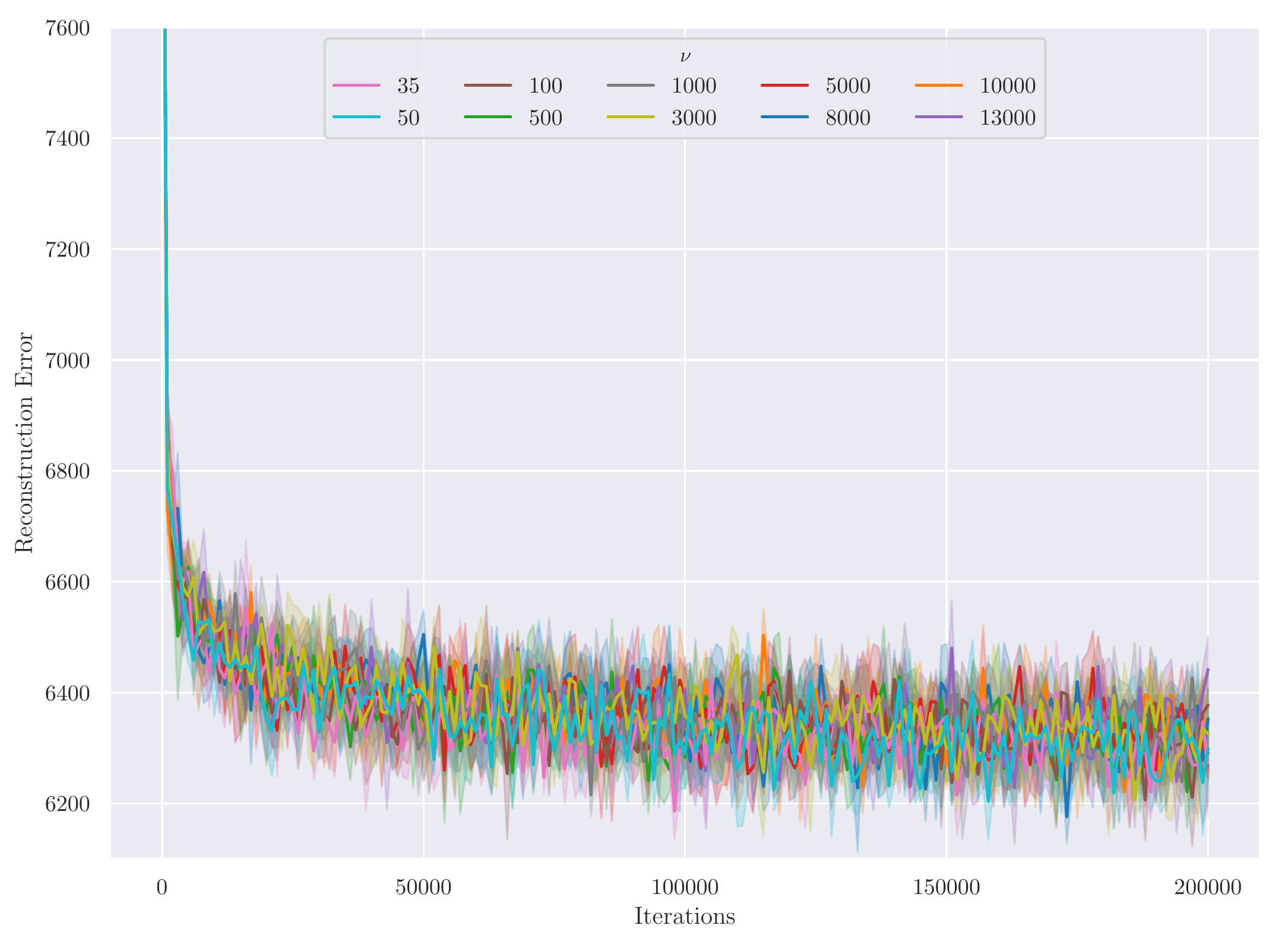}}
	\subfloat{\includegraphics[width=0.32\textwidth]{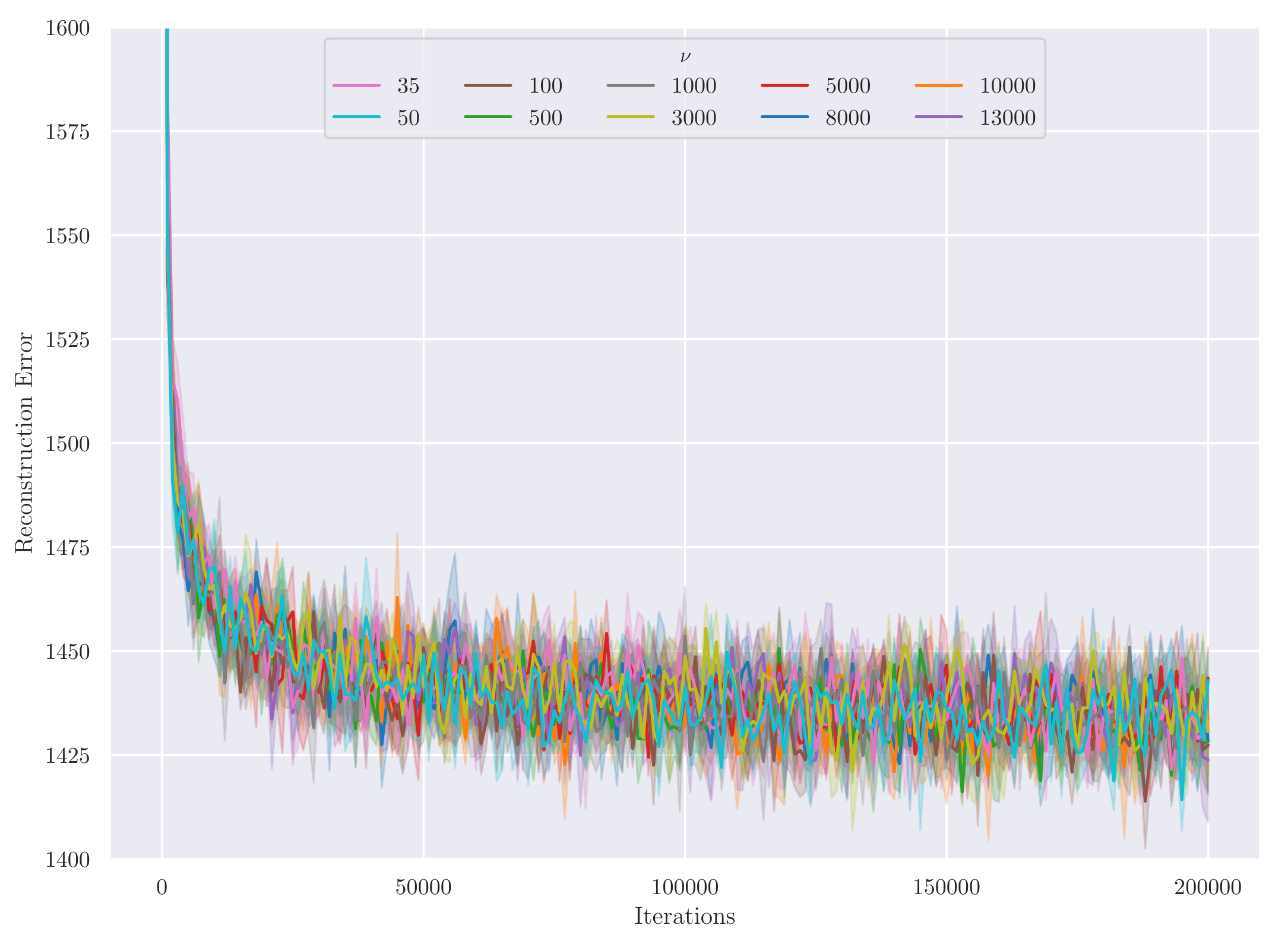}}    \subfloat{\includegraphics[width=0.32\textwidth]{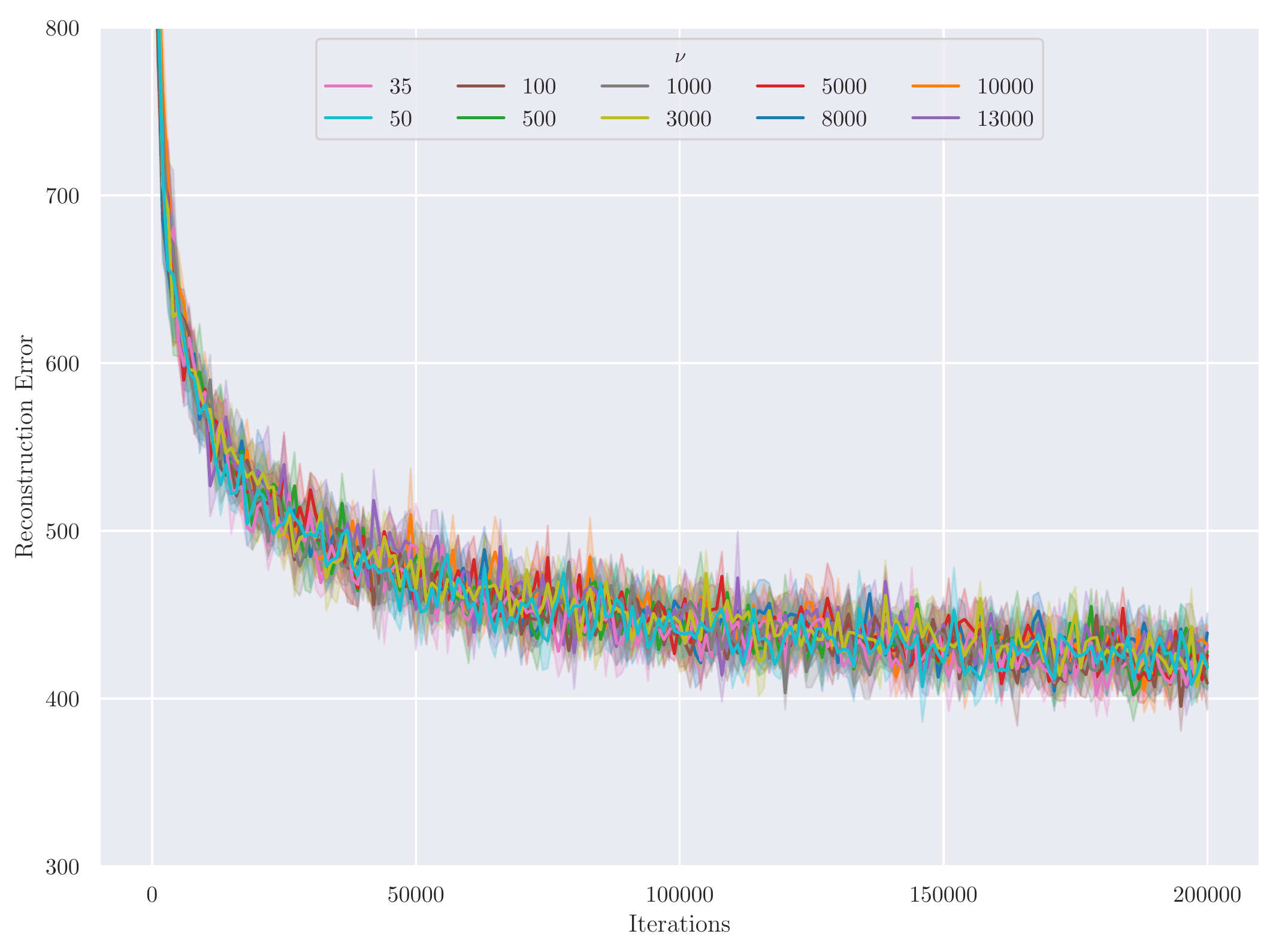}}
	\caption{Variation of reconstruction error with iterations for different values of $\nu$ in CHyVAE on CelebA (left), 3DFaces (center), and 3DChairs (right) dataset averaged over 5 random restarts}
	\label{fig:recons_chyvae}
\end{figure*}
\begin{figure}
	\centering
	\includegraphics[width=0.9\columnwidth]{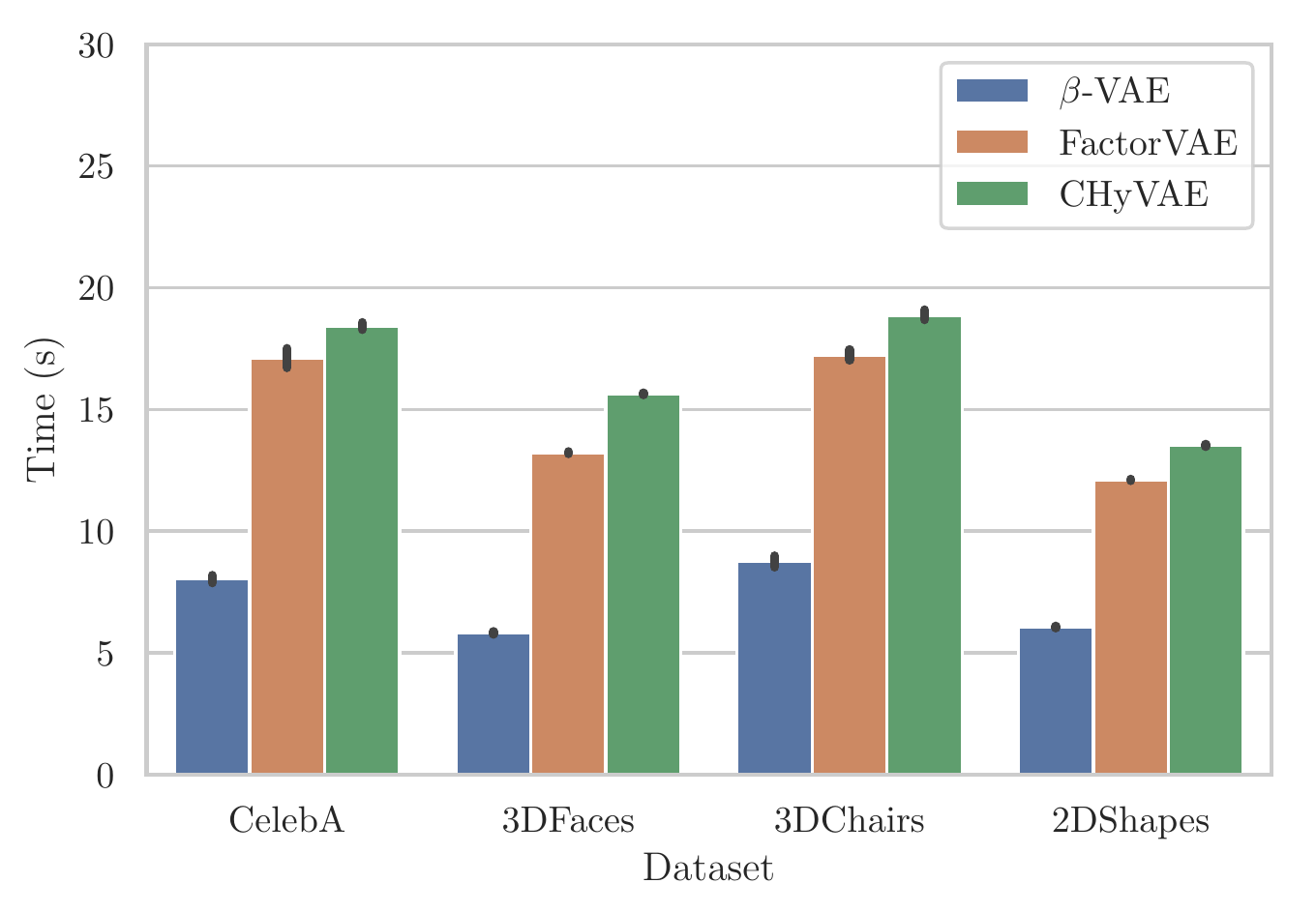}
	\caption{Average runtime for $\beta$-VAE, FactorVAE, and CHyVAE on different datasets}
	\label{fig:runtime}			
\end{figure}
\begin{figure}
	\centering
	\includegraphics[width=.9\columnwidth]{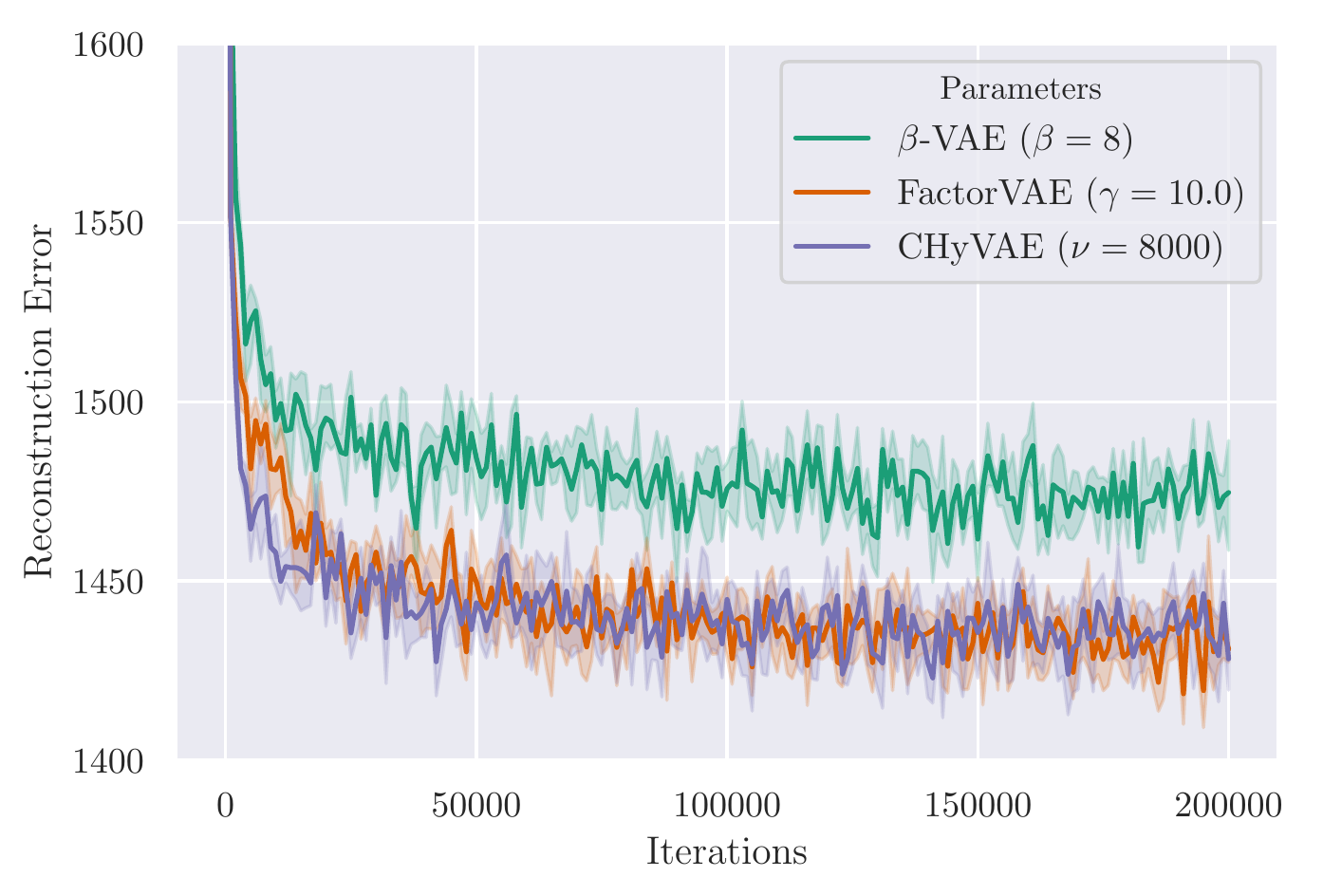}
	\caption{Variation of reconstruction error with iterations for $\beta$-VAE, FactorVAE, and CHyVAE on 3DFaces dataset averaged over 5 random restarts}
	\label{fig:3dfaces_recons}
\end{figure}
\begin{figure}
	\centering
	\includegraphics[width=0.9\columnwidth]{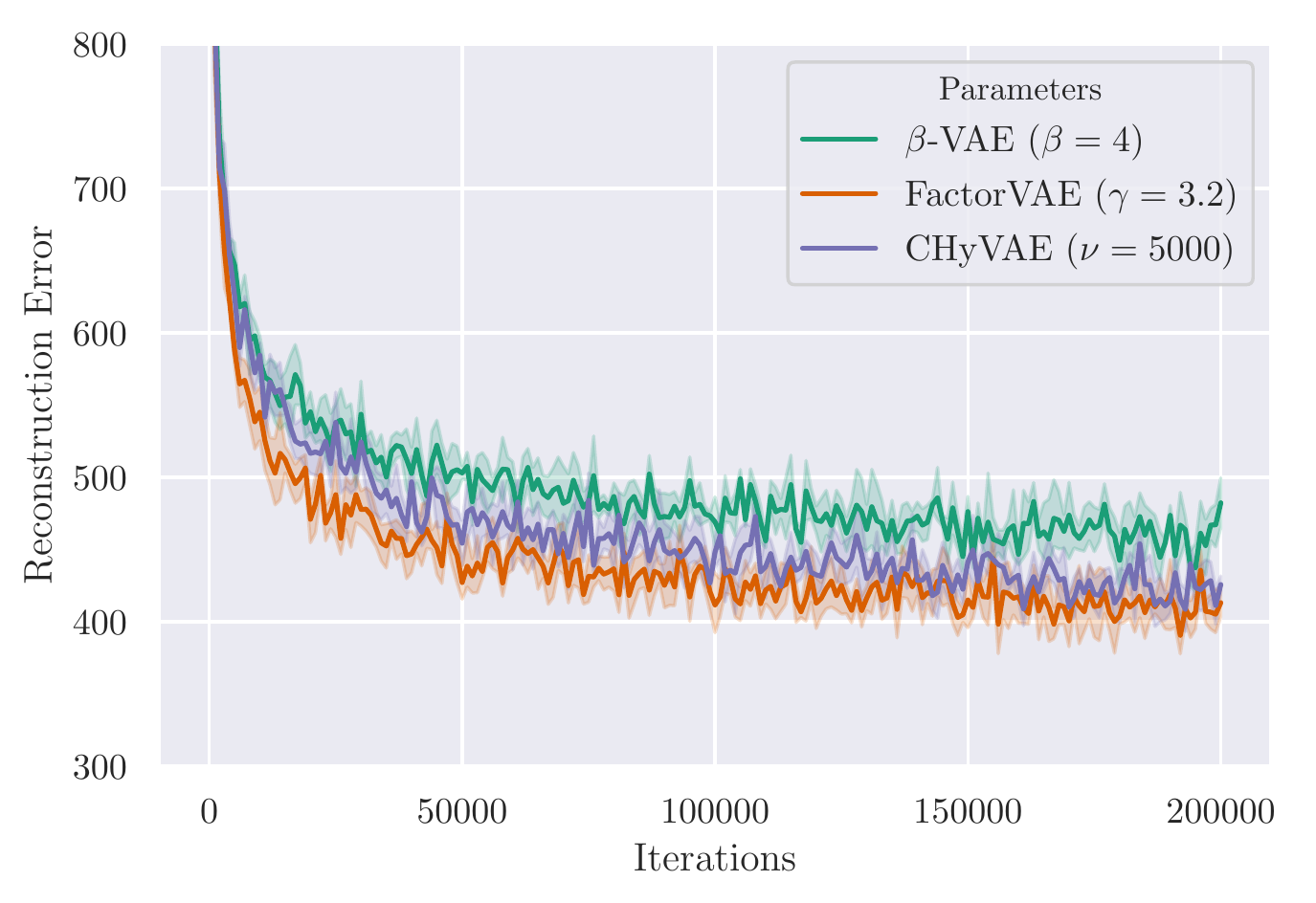}
	\caption{Variation of reconstruction error with iterations for $\beta$-VAE, FactorVAE, and CHyVAE on 3DChairs dataset averaged over 5 random restarts}
	\label{fig:3dchairs_recons}			
\end{figure}
\begin{figure}
	\centering
	\subfloat{\includegraphics[width=.6\columnwidth]{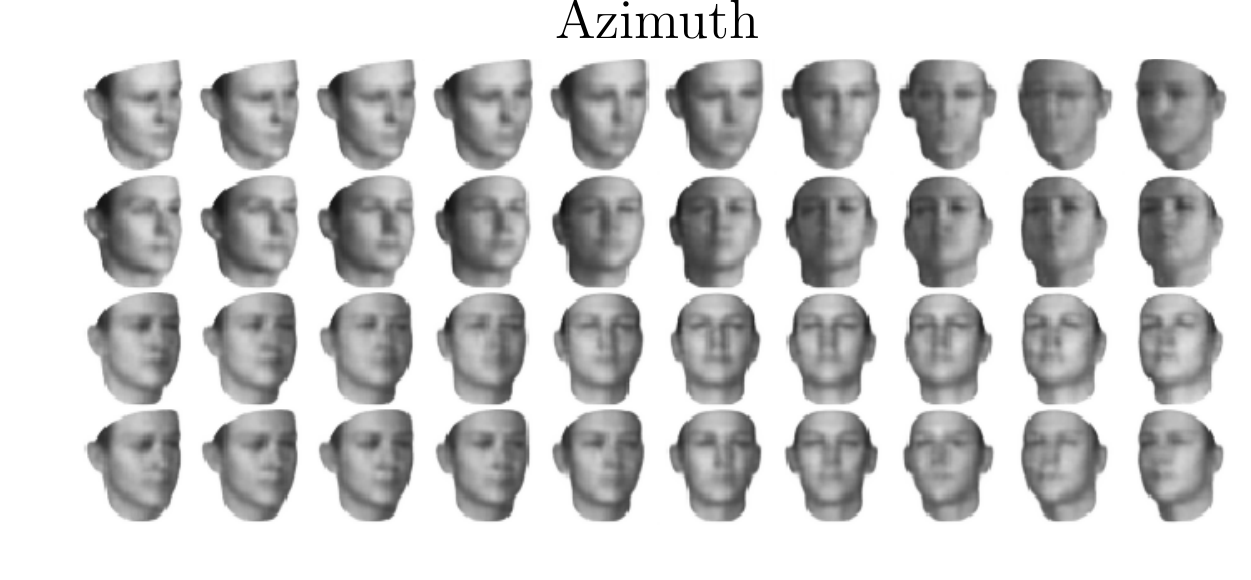}}\\
	\subfloat{\includegraphics[width=.6\columnwidth]{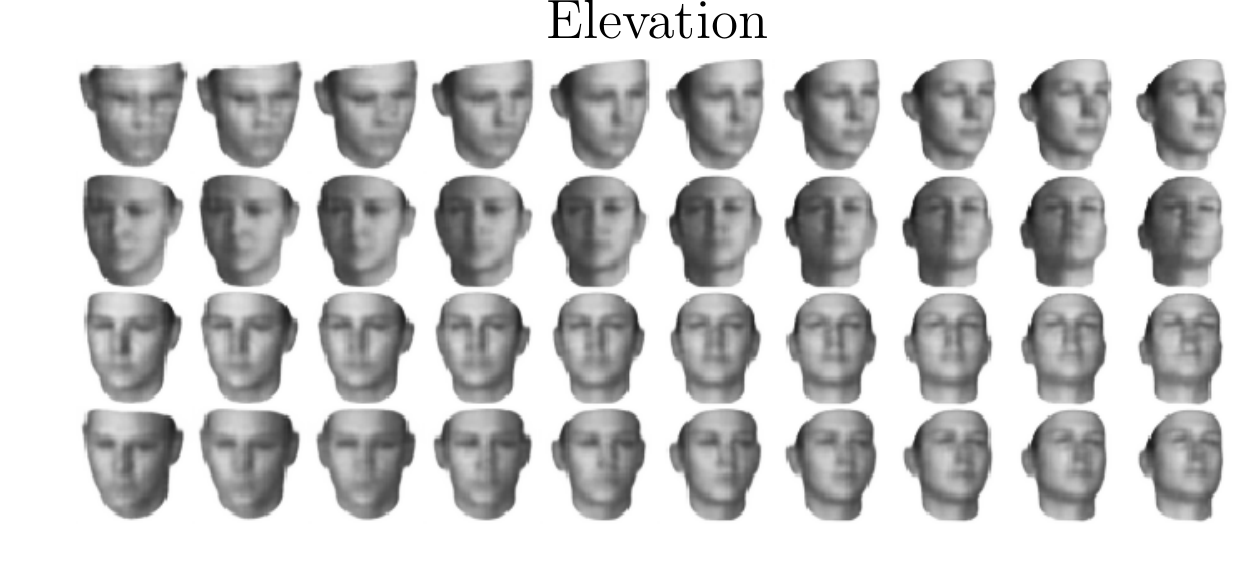}}\\
	\subfloat{\includegraphics[width=.6\columnwidth]{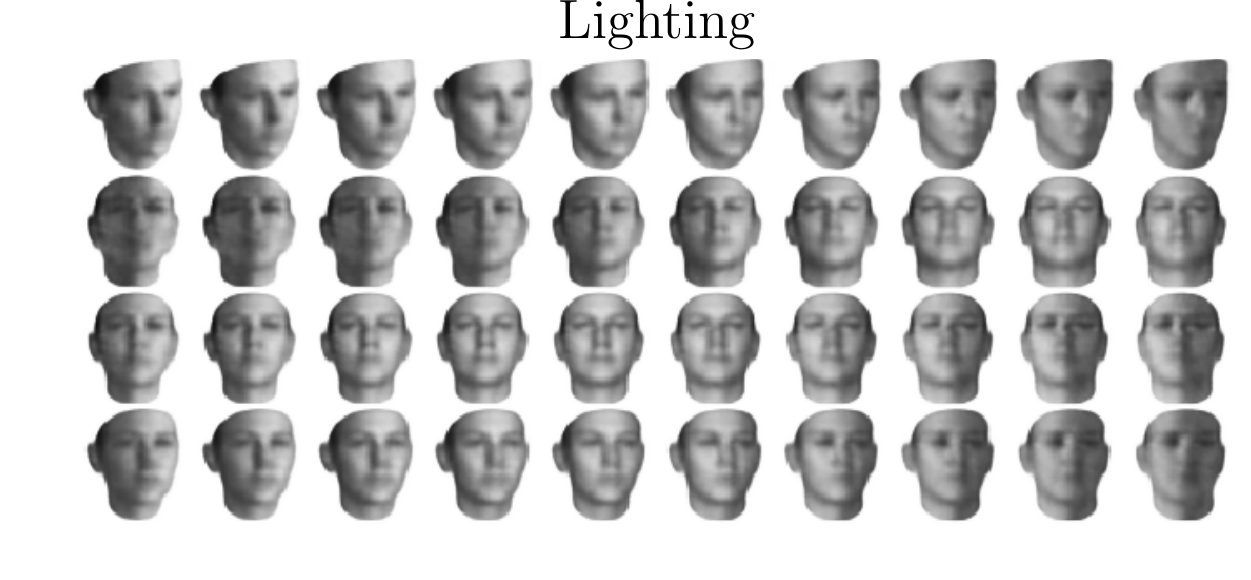}}
	\caption{Latent traversals for CHyVAE on 3DFaces}
	\label{fig:3dfaces_extras}
\end{figure}

\begin{figure*}
	\centering
	\subfloat{\includegraphics[width=.33\textwidth]{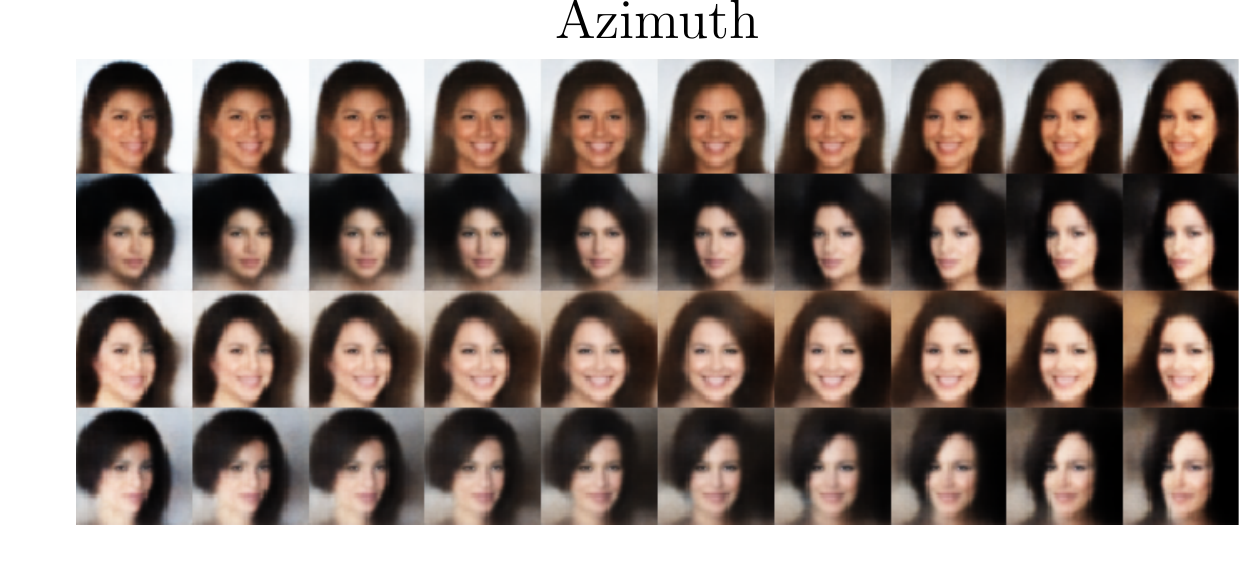}}
	\subfloat{\includegraphics[width=.33\textwidth]{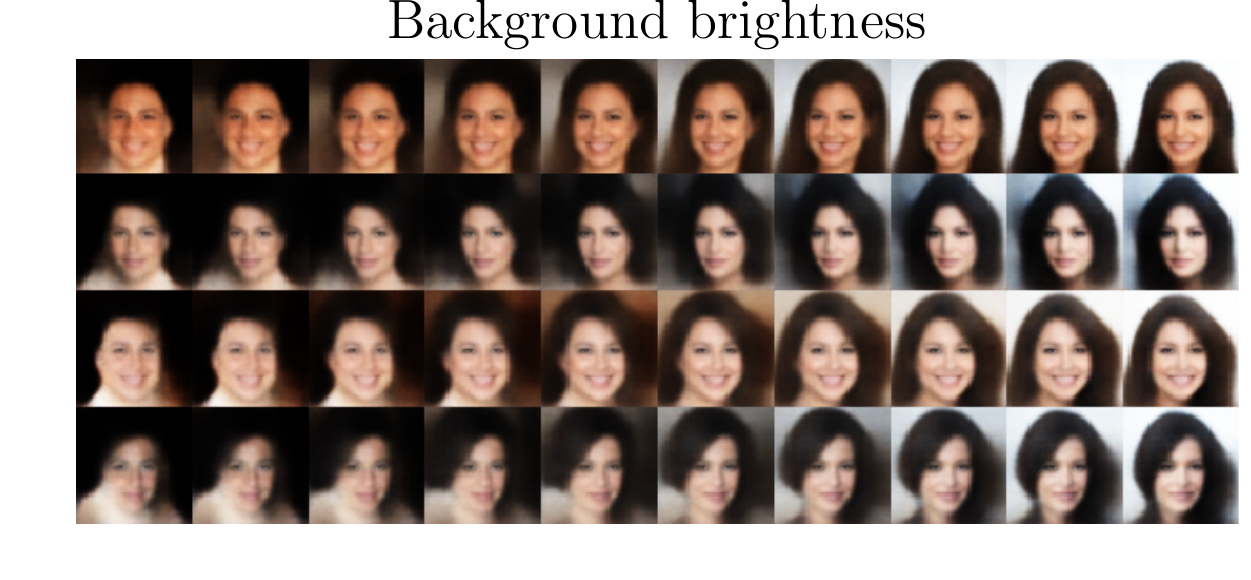}}
	\subfloat{\includegraphics[width=.33\textwidth]{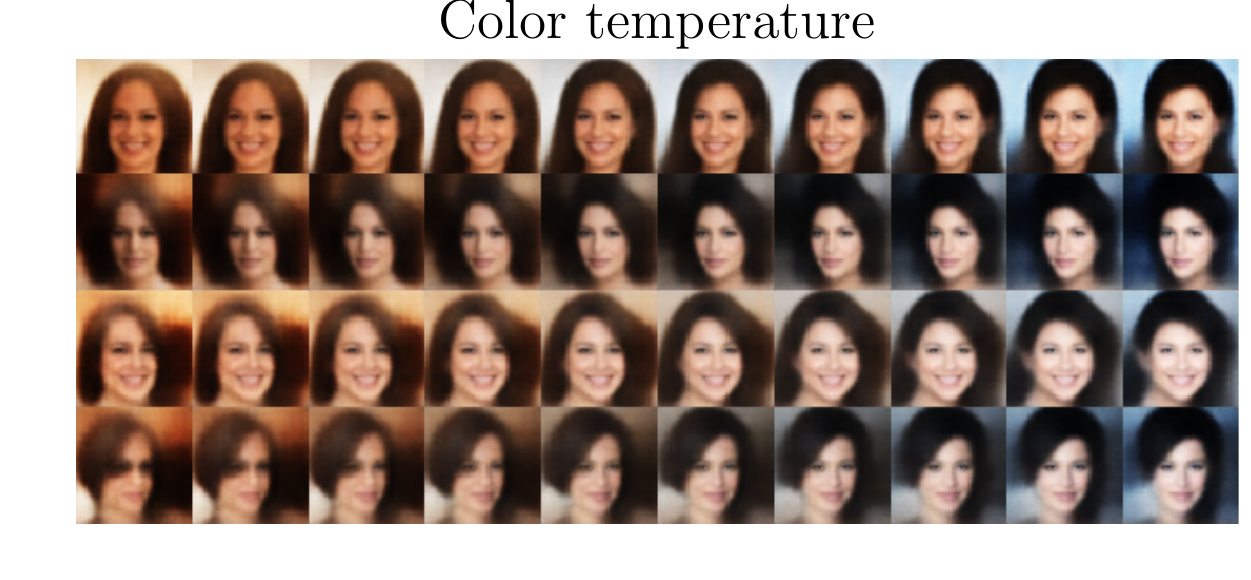}}\\
	\subfloat{\includegraphics[width=.33\textwidth]{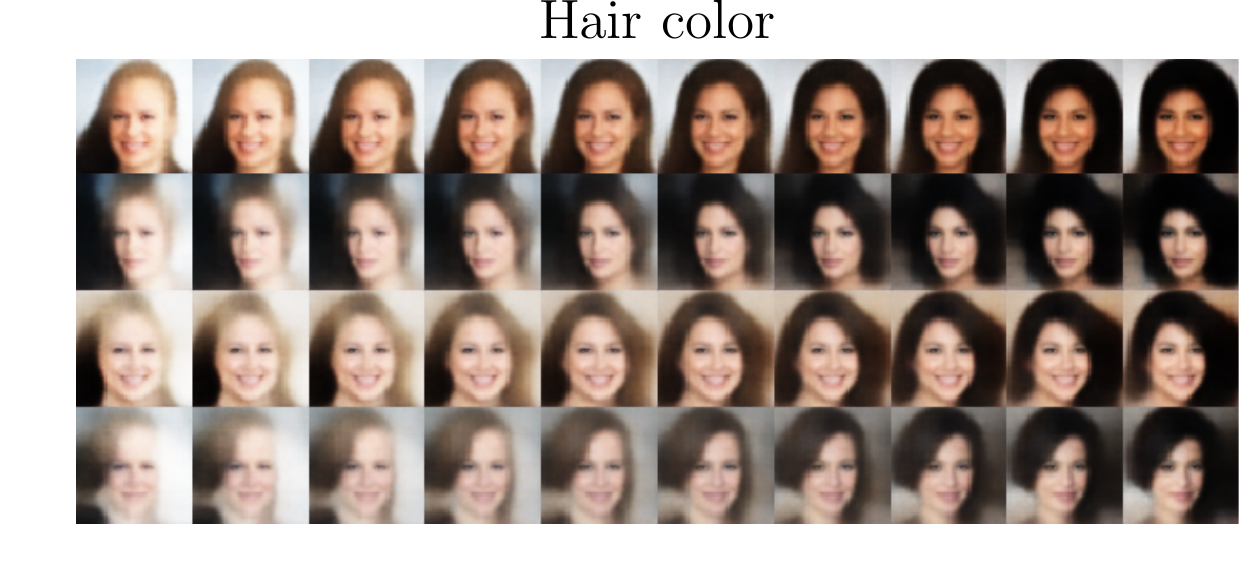}}
	\subfloat{\includegraphics[width=.33\textwidth]{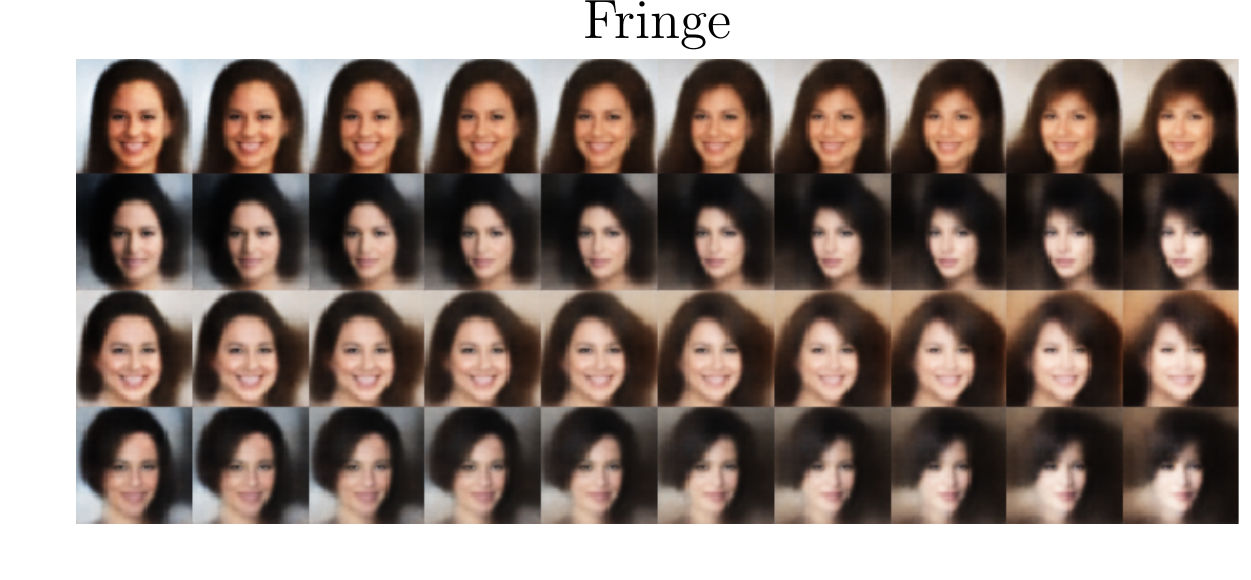}}
	\subfloat{\includegraphics[width=.33\textwidth]{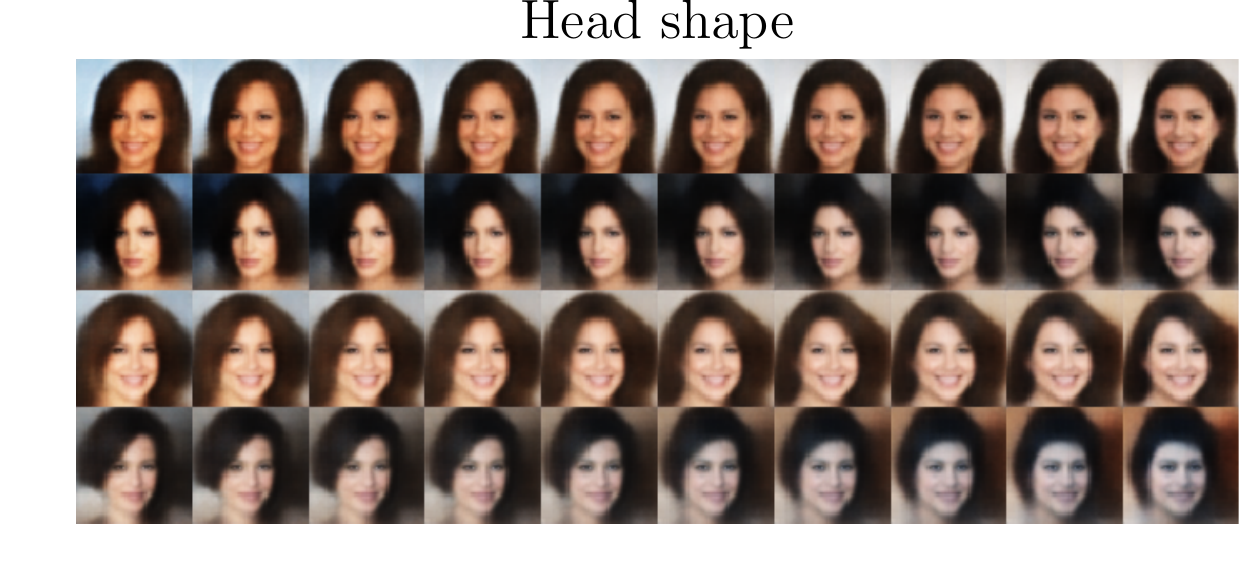}}\\
	\subfloat{\includegraphics[width=.33\textwidth]{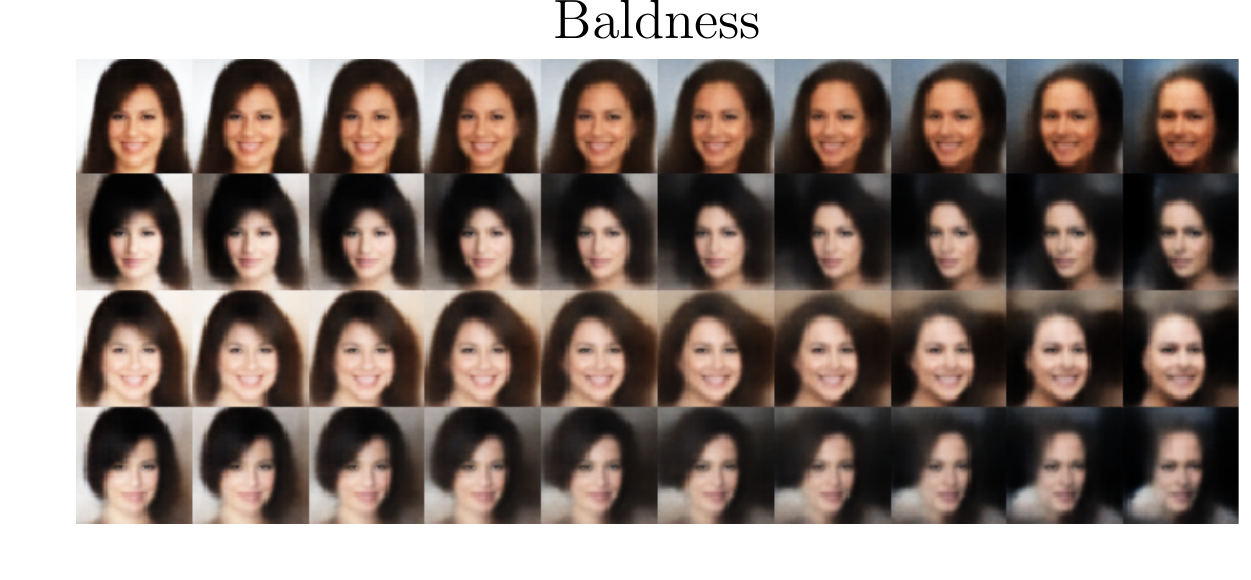}}
	\subfloat{\includegraphics[width=.33\textwidth]{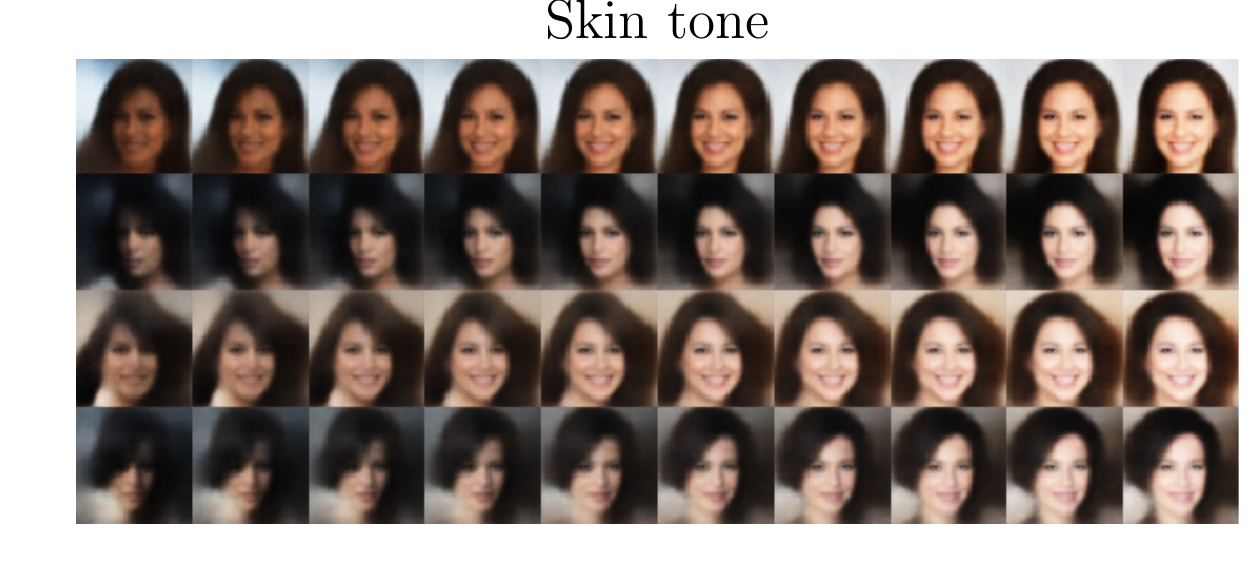}}
	\subfloat{\includegraphics[width=.33\textwidth]{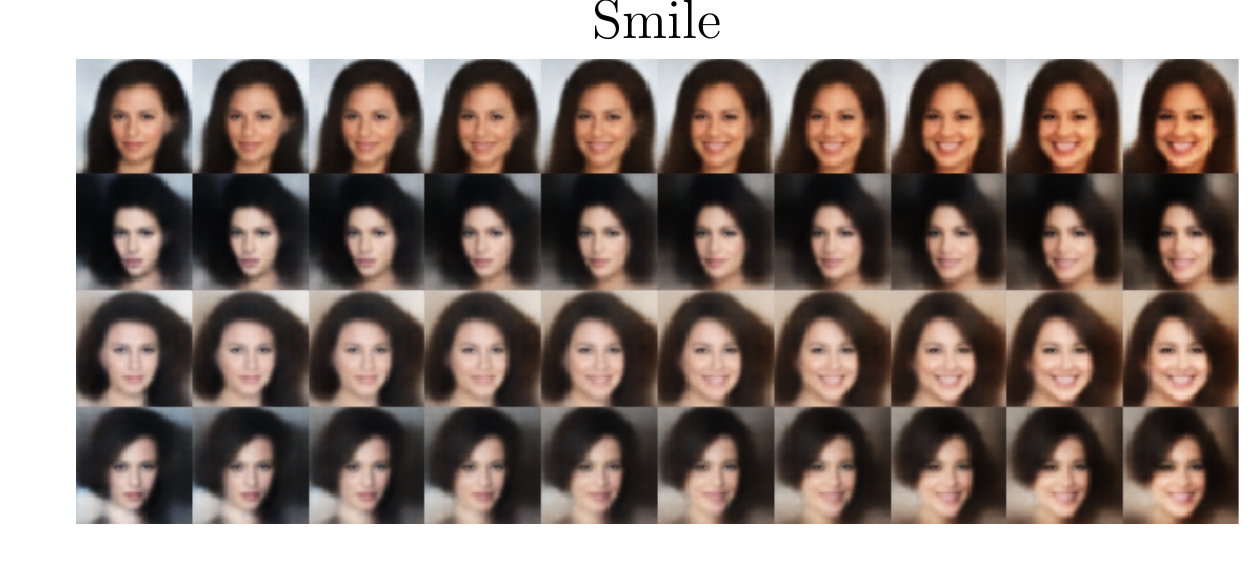}}
	\caption{Latent traversals for CHyVAE on CelebA}
	\label{fig:celeba_extras}
\end{figure*}

\begin{figure*}
	\centering
	\subfloat{\includegraphics[width=.5\textwidth]{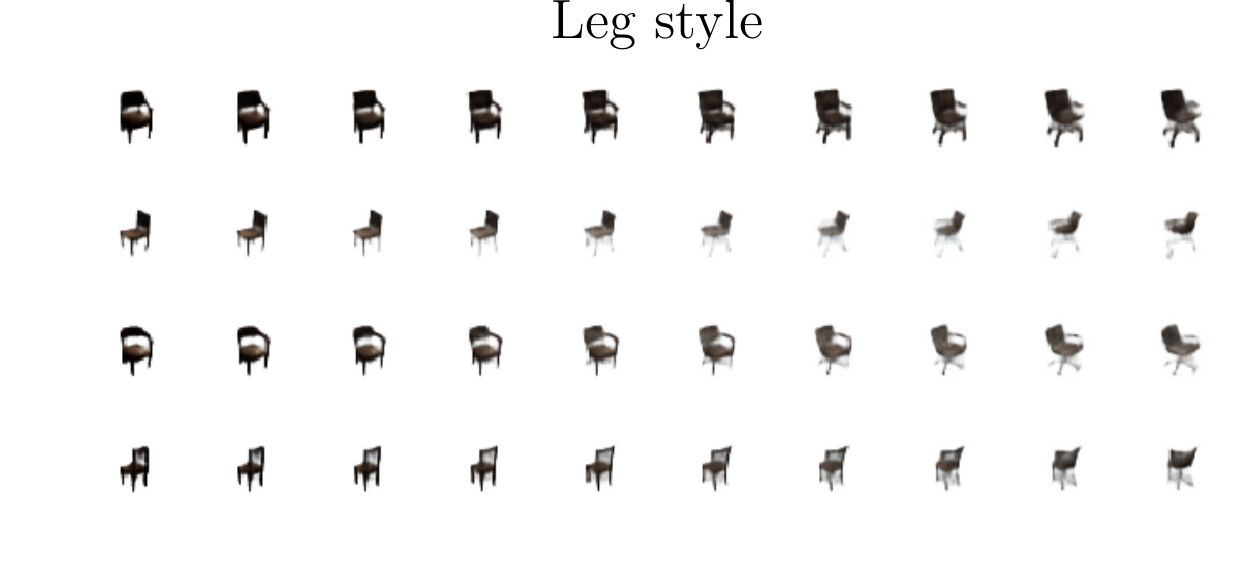}}
	\subfloat{\includegraphics[width=.5\textwidth]{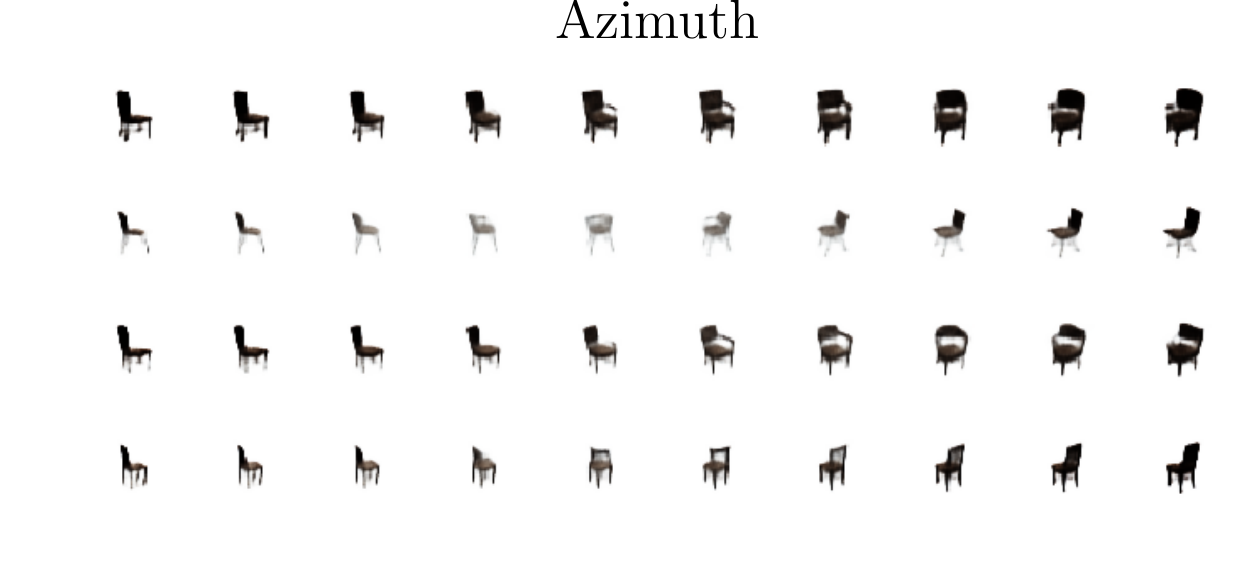}}\\
	\subfloat{\includegraphics[width=.5\textwidth]{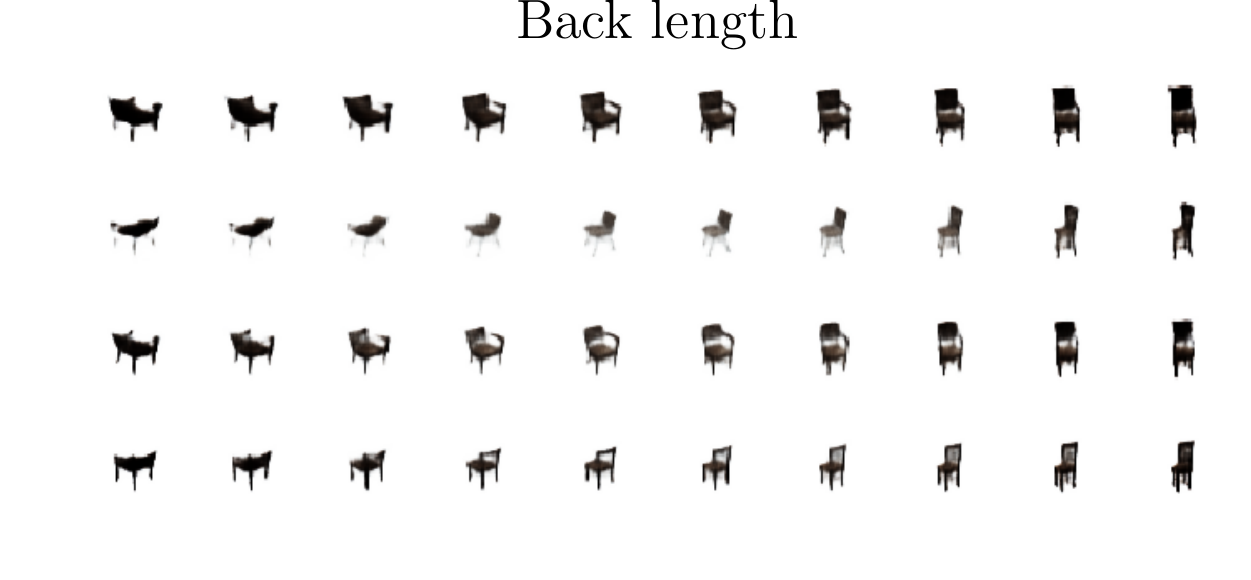}}
	\subfloat{\includegraphics[width=.5\textwidth]{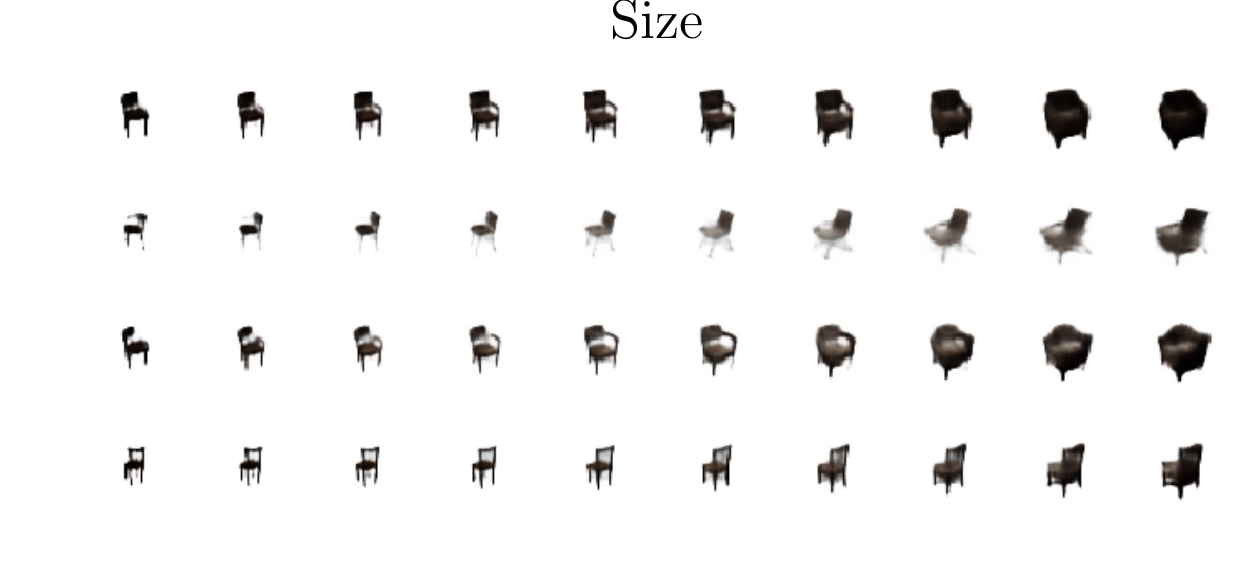}}
	\caption{Latent traversals for CHyVAE on 3DChairs}
	\label{fig:3dchairs_extras}
\end{figure*}

\subsection{Other Results}
Figures \ref{fig:3dfaces_recons} and \ref{fig:3dchairs_recons} show the variation of reconstruction error for $\beta$-VAE, FactorVAE, and CHyVAE with training iterations for 3DFaces and 3DChairs. Figure \ref{fig:recons_chyvae} shows the variation of reconstruction error with the number of iterations for different values of the hyperparameter $\nu$ in CHyVAE on CelebA, 3DFaces, and 3DChairs. Figure \ref{fig:2dshapes_chyvae_inter} shows the latent traversals on 2DShapes dataset. Figure \ref{fig:runtime} shows the average time taken by each model for training on 1000 minibatches of batch-size 50. FactorVAE and CHyVAE require more computation time owing to the discriminator network and matrix operations in the graph respectively. Figs. \ref{fig:3dfaces_extras}, \ref{fig:celeba_extras}, and \ref{fig:3dchairs_extras} show additional latent traversals for CHyVAE. \footnote{Visit \url{https://github.com/crslab/CHyVAE} for more qualitative results.}

\onecolumn
\subsection{Derivation of the ELBO decomposition}
The ELBO in Eq. (\ref{eq:elbo}), when averaged over the data-points, can be written as
\begin{align}
\mathcal{L}_{\mathrm{ELBO}}^\text{avg.} &= \mathbb{E}_{q(\mathbf{x})}\left[\mathbb{E}_{q(\mathbf{z},\bm{\Sigma}|\mathbf{x})}\left[\frac{p(\mathbf{x}|\mathbf{z})p(\mathbf{z},\bm{\Sigma})}{q(\mathbf{z},\bm{\Sigma}|\mathbf{x})}\right]\right]\\
&=\mathbb{E}_{q(\mathbf{z},\bm{\Sigma},\mathbf{x})}\left[\frac{p(\mathbf{x}|\mathbf{z})p(\mathbf{z},\bm{\Sigma})q(\mathbf{z},\bm{\Sigma})}{q(\mathbf{z},\bm{\Sigma}|\mathbf{x})q(\mathbf{z},\bm{\Sigma})}\right]\\
&=\mathbb{E}_{q(\mathbf{z},\bm{\Sigma},\mathbf{x})}\left[p(\mathbf{x}|\mathbf{z})\right] + \mathbb{E}_{q(\mathbf{z},\bm{\Sigma},\mathbf{x})}\left[\frac{q(\mathbf{z},\bm{\Sigma})}{q(\mathbf{z},\bm{\Sigma}|\mathbf{x})}\right] + \mathbb{E}_{q(\mathbf{z},\bm{\Sigma},\mathbf{x})}\left[\frac{p(\mathbf{z},\bm{\Sigma})}{q(\mathbf{z},\bm{\Sigma})}\right]\\
&=\mathbb{E}_{q(\mathbf{z},\bm{\Sigma},\mathbf{x})}\left[p(\mathbf{x}|\mathbf{z})\right] + \mathbb{E}_{q(\mathbf{z},\bm{\Sigma},\mathbf{x})}\left[\frac{q(\bm{\Sigma}|\mathbf{z})q(\mathbf{z})}{q(\mathbf{z}|\mathbf{x})q(\bm{\Sigma}|\mathbf{z})}\right] + \mathbb{E}_{q(\mathbf{z},\bm{\Sigma},\mathbf{x})}\left[\frac{p(\mathbf{z},\bm{\Sigma})}{q(\mathbf{z},\bm{\Sigma})}\right]\\
&=\mathbb{E}_{q(\mathbf{z},\mathbf{x})}\left[p(\mathbf{x}|\mathbf{z})\right] + \mathbb{E}_{q(\mathbf{z},\mathbf{x})}\left[\frac{q(\mathbf{x})q(\mathbf{z})}{q(\mathbf{z},\mathbf{x})}\right] + \mathbb{E}_{q(\mathbf{z})}\left[\frac{p(\mathbf{z})}{q(\mathbf{z})}\right] + \mathbb{E}_{q(\mathbf{z},\bm{\Sigma})}\left[\frac{p(\bm{\Sigma}|\mathbf{z})}{q(\bm{\Sigma}|\mathbf{z})}\right]\\
&=\mathbb{E}_{q(\mathbf{x})}\left[\mathbb{E}_{q(\mathbf{z}|\mathbf{x})}\left[p(\mathbf{x}|\mathbf{z})\right]\right] - I(\mathbf{x};\mathbf{z}) - D_{\mathrm{KL}}\left(q(\mathbf{z})\|p(\mathbf{z})\right) - \mathbb{E}_{q(\mathbf{z})}\left[D_{\mathrm{KL}}\left(q(\bm{\Sigma}|\mathbf{z})\|p(\bm{\Sigma}|\mathbf{z})\right)\right]\\
&= \underset{\textnormal{(average reconstruction)}}{\left[\frac{1}{N}\sum_{n=1}^N\mathbb{E}_{q(\mathbf{z}|\mathbf{x})}\left[\log p(\mathbf{x}|\mathbf{z})\right]\right]} - \underset{\textnormal{(index-code MI)}}{I(\mathbf{x};\mathbf{z})}
- \underset{\textnormal{(marginal KL to prior)}}{D_{\mathrm{KL}}(q(\mathbf{z})\|p(\mathbf{z}))} - \underset{\textnormal{(covariance penalty)}}{\mathbb{E}_{q(\mathbf{z})}\left[D_{\mathrm{KL}}(q(\bm{\Sigma}|\mathbf{z})\|p(\bm{\Sigma}|\mathbf{z}))\right]}
\end{align}
\subsection{Derivation of the closed-form expression for the ELBO}
\label{sec:elbofullderivation}
From Eq. (13) in the paper, we have the following expression of $\mathcal{L}_{\mathrm{ELBO}}$.
\begin{align}
\mathcal{L}_{\mathrm{ELBO}} &= \underset{1}{\underbrace{\mathbb{E}_{q(\mathbf{z}|\mathbf{x})}\left[\log p(\mathbf{x}|\mathbf{z})\right]}} -\underset{2}{\underbrace{\mathbb{E}_{p(\bm{\Sigma}|\mathbf{z})}\left[D_{\mathrm{KL}}(\mathcal{N}(\tilde{\bm{\mu}}, \tilde{\bm{\Sigma}})||\mathcal{N}(\bm{0}, \bm{\Sigma}))\right]}} -\underset{3}{\underbrace{\mathbb{E}_{q(\mathbf{z}|\mathbf{x})}\left[D_{\mathrm{KL}}(\mathcal{W}_p^{-1}(\bm{\Phi}, \lambda)||\mathcal{W}_p^{-1}(\bm{\Psi}, \nu))\right]}}
\label{eq:elbosupp}
\end{align}
Henceforth, the derivation uses the following known results.
\begin{itemize}
	\item For a random variable $\mathbf{X} \sim \mathcal{W}_p^{-1}(\bm{\Psi},\nu)$, 
	\begin{align}
	\mathbb{E}\left[\log|\mathbf{X}|\right] &= \log\frac{|\bm{\Psi}|}{2^p} - \psi_{p}\left(\frac{\nu}{2}\right)\label{eq:supexplogx}\\
	\mathbb{E}\left[\mathbf{X}^{-1}\right] &= \nu\bm{\Psi}^{-1}
	\label{eq:supexpinvx}
	\end{align}
	\item For two multivariate normal distributions $p = \mathcal{N}(\bm{\mu}_0, \bm{\Sigma}_0)$ and $q = \mathcal{N}(\bm{\mu}_1, \bm{\Sigma}_1)$,
	\begin{align}
	D_{\mathrm{KL}}(q||p) &= \frac{1}{2} \Bigg(\log \frac{|\bm{\Sigma}_0|}{|\bm{\Sigma}_1|} - p + \Tr(\bm{\Sigma}_0^{-1}\bm{\Sigma}_1) + (\bm{\mu}_0 - \bm{\mu}_1)^{\top}\bm{\Sigma}_0^{-1}(\bm{\mu}_0 - \bm{\mu}_1)\Bigg)
	\label{eq:supklmvn}
	\end{align}
	\item For two Inverse-Wishart distributions $p = \mathcal{W}_p^{-1}(\bm{\Psi}_0,\nu_0)$ and $q = \mathcal{W}_p^{-1}(\bm{\Psi}_1,\nu_1)$,
	\begin{align}
	D_{\mathrm{KL}}(q||p) &= -\frac{\nu_0}{2}\log|\bm{\Psi}_0\bm{\Psi}_1^{-1}| + \frac{\nu_1}{2}\left(\Tr(\bm{\Psi}_0\bm{\Psi}_1^{-1})-p\right) +\log\frac{\Gamma_p\left(\frac{\nu_0}{2}\right)}{\Gamma_p\left(\frac{\nu_1}{2}\right)} + \frac{\nu_1-\nu_0}{2}\psi_{p}\left(\frac{\nu_1}{2}\right)
	\label{eq:supklwishart}
	\end{align}
	where $\Tr$ is the matrix trace function and $\psi_{p}$ is the multivariate digamma function.
\end{itemize}

\noindent For a Bernoulli distribution with mean $\hat{\mathbf{x}}$, a single sample estimate of the first term in Eq. (\ref{eq:elbosupp}) can be written as
\begin{align}
\mathbb{E}_{q(\mathbf{z}|\mathbf{x})}\left[\log p(\mathbf{x}|\mathbf{z})\right] = \sum_{j=1}^{D}\left(\mathbf{x}_i|_j\log \hat{\mathbf{x}}_i|_j + (1-\mathbf{x}_i|_j)\log(1 - \hat{\mathbf{x}}_i|_j)\right)\label{eq:bernlike}
\end{align}
where $D$ is the dimensionality of $\mathbf{x}_i$.
\\\\
The second term in Eq. (\ref{eq:elbosupp}) can be expanded as follows.
\begin{align}
\mathbb{E}_{p(\bm{\Sigma}|\mathbf{z})}\left[D_{KL}(\mathcal{N}(\mathbf{z};\tilde{\bm{\mu}}, \tilde{\bm{\Sigma}})||\mathcal{N}(\mathbf{z};\bm{0}, \bm{\Sigma}))\right] &= \frac{1}{2}\mathbb{E}_{p(\bm{\Sigma}|\mathbf{z})}\left[\log \frac{|\bm{\Sigma}|}{|\tilde{\bm{\Sigma}}|} - p + \Tr(\bm{\Sigma}^{-1}\tilde{\bm{\Sigma}}) +  \tilde{\bm{\mu}}^{\top}\bm{\Sigma}^{-1}\tilde{\bm{\mu}}\right]\label{eq:klnml}\\
&=\frac{1}{2}\mathbb{E}_{p(\bm{\Sigma}|\mathbf{z})}\left[\log \frac{|\bm{\Sigma}|}{|\tilde{\bm{\Sigma}}|} - p + \Tr(\bm{\Sigma}^{-1}(\tilde{\bm{\Sigma}} + \tilde{\bm{\mu}}\tilde{\bm{\mu}}^\top)) \right]\label{eq:cycletrace}\\\nonumber
&=\frac{1}{2}\left[- p - \log {|\tilde{\bm{\Sigma}}|}\right]\\
&\quad+ \frac{1}{2}\mathbb{E}_{p(\bm{\Sigma}|\mathbf{z})}\left[\log {|\bm{\Sigma}|} + \Tr(\bm{\Sigma}^{-1}(\tilde{\bm{\Sigma}} + \tilde{\bm{\mu}}\tilde{\bm{\mu}}^\top)) \right]\label{eq:constsep}\\\nonumber
&=\frac{1}{2}\left[- p - \log {|\tilde{\bm{\Sigma}}|} + \log\frac{|\bm{\Phi}|}{2^p} - \psi_{p}\left(\frac{\lambda}{2}\right)\right]\\
&\quad+ \frac{1}{2}\Tr\left(\mathbb{E}_{p(\bm{\Sigma}|\mathbf{z})}\left[\bm{\Sigma}^{-1}\right](\tilde{\bm{\Sigma}} + \tilde{\bm{\mu}}\tilde{\bm{\mu}}^\top)\right)\label{eq:explg}\\\nonumber
&=\frac{1}{2}\left[- p - \log {|\tilde{\bm{\Sigma}}|} + \log\frac{|\bm{\Phi}|}{2^p} - \psi_{p}\left(\frac{\lambda}{2}\right)\right]\\
&\quad+ \frac{1}{2}\Tr\left(\lambda\bm{\Phi}^{-1}(\tilde{\bm{\Sigma}} + \tilde{\bm{\mu}}\tilde{\bm{\mu}}^\top)\right)\label{eq:expinvw}\\\nonumber
&=\frac{1}{2}\left[ - \log {|\tilde{\bm{\Sigma}}|} + \log{|\bm{\Phi}|} + \Tr\left(\lambda\bm{\Phi}^{-1}(\tilde{\bm{\Sigma}} + \tilde{\bm{\mu}}\tilde{\bm{\mu}}^\top)\right)\right]\\
&\quad+ \underset{\textnormal{constants}}{\underbrace{\frac{1}{2}\left[-p -\log{2^p}- \psi_{p}\left(\frac{\lambda}{2}\right)\right]}}\label{eq:term2}
\end{align}
where Eq. (\ref{eq:klnml}) uses Eq. (\ref{eq:supklmvn}), Eq. (\ref{eq:cycletrace}) uses the properties of the $\mathrm{trace}$ function exploiting the fact that $\tilde{\bm{\mu}}^{\top}\bm{\Sigma}^{-1}\tilde{\bm{\mu}}$ can be written as $\Tr(\tilde{\bm{\mu}}^{\top}\bm{\Sigma}^{-1}\tilde{\bm{\mu}})$, Eq. (\ref{eq:constsep}) separates terms constant with respect to the expectation, Eq. (\ref{eq:explg}) uses Eq. (\ref{eq:supexplogx}) and the linearity of $\Tr$ and $\mathbb{E}$, Eq. (\ref{eq:expinvw}) uses Eq. (\ref{eq:supexpinvx}).
\\\\
The KL-divergence in the third term in Eq. (\ref{eq:elbosupp}) can be expanded as
\begin{align}
\nonumber D_{\mathrm{KL}}(\mathcal{W}_p^{-1}(\bm{\Phi}, \lambda)||\mathcal{W}_p^{-1}(\bm{\Psi}, \nu)) &= -\frac{\nu}{2}\log|\bm{\Psi}\bm{\Phi}^{-1}| + \frac{\lambda}{2}\left(\Tr(\bm{\Psi}\bm{\Phi}^{-1})-p\right)\\
&\quad+ \log\frac{\Gamma_p\left(\frac{\nu}{2}\right)}{\Gamma_p\left(\frac{\lambda}{2}\right)} + \frac{\lambda-\nu}{2}\psi_{p}\left(\frac{\lambda}{2}\right)\label{eq:wishstep1}\\\nonumber
&= -\frac{\nu}{2}\log|\bm{\Psi}||\bm{\Phi}^{-1}| + \frac{\lambda}{2}\left(\Tr(\bm{\Psi}\bm{\Phi}^{-1})\right)\\
&\quad- \frac{p\lambda}{2}+ \log\frac{\Gamma_p\left(\frac{\nu}{2}\right)}{\Gamma_p\left(\frac{\lambda}{2}\right)} + \frac{\lambda-\nu}{2}\psi_{p}\left(\frac{\lambda}{2}\right)\label{eq:wishstep2}\\\nonumber
&= -\frac{\nu}{2}\log|\bm{\Psi}|+\frac{\nu}{2}\log|\bm{\Phi}| + \frac{\lambda}{2}\left(\Tr(\bm{\Psi}\bm{\Phi}^{-1})\right)\\
&\quad- \frac{p\lambda}{2}+ \log\frac{\Gamma_p\left(\frac{\nu}{2}\right)}{\Gamma_p\left(\frac{\lambda}{2}\right)} + \frac{\lambda-\nu}{2}\psi_{p}\left(\frac{\lambda}{2}\right)\label{eq:wishstep3}\\\nonumber
&= \frac{\nu}{2}\log|\bm{\Phi}| + \frac{\lambda}{2}\left(\Tr(\bm{\Psi}\bm{\Phi}^{-1})\right)\\
&\quad-\underset{\textnormal{constants}}{\underbrace{\frac{\nu}{2}\log|\bm{\Psi}| - \frac{p(\lambda)}{2}+ \log\frac{\Gamma_p\left(\frac{\nu}{2}\right)}{\Gamma_p\left(\frac{\lambda}{2}\right)} + \frac{\lambda-\nu}{2}\psi_{p}\left(\frac{\lambda}{2}\right)}}\label{eq:wishstep4}\\
&= \frac{\nu}{2}\log|\bm{\Phi}| + \frac{\lambda}{2}\left(\Tr(\bm{\Psi}\bm{\Phi}^{-1})\right)\label{eq:term3}
\end{align}
where Eq. (\ref{eq:wishstep1}) uses Eq. (\ref{eq:supklwishart}), Eq. (\ref{eq:wishstep2}) and (\ref{eq:wishstep3}) use properties of determinants, and Eq. (\ref{eq:wishstep4}) separates constant terms.

Eq. (\ref{eq:term2}) and (\ref{eq:term3}) can be combined as follows.

\begin{align}
\mathcal{L}_{\mathrm{ELBO}} &= \mathbb{E}_{q(\mathbf{z}|\mathbf{x})}\left[\log p(\mathbf{x}|\mathbf{z})\right] - \frac{1}{2}\left[ - \log {|\tilde{\bm{\Sigma}}|} + \log{|\bm{\Phi}|} + \Tr\left(\lambda\bm{\Phi}^{-1}(\tilde{\bm{\Sigma}} + \tilde{\bm{\mu}}\tilde{\bm{\mu}}^\top)\right)\right] - \left[\frac{\nu}{2}\log|\bm{\Phi}| + \frac{\lambda}{2}\left(\Tr(\bm{\Psi}\bm{\Phi}^{-1})\right)\right]\\
\mathcal{L}_{\mathrm{ELBO}} &= \mathbb{E}_{q(\mathbf{z}|\mathbf{x})}\left[\log p(\mathbf{x}|\mathbf{z})\right] + \frac{1}{2} \log {|\tilde{\bm{\Sigma}}|} - \frac{1}{2} \log{|\bm{\Phi}|} - \frac{1}{2} \Tr\left(\lambda\bm{\Phi}^{-1}(\tilde{\bm{\Sigma}} + \tilde{\bm{\mu}}\tilde{\bm{\mu}}^\top)\right) - \frac{\nu}{2}\log|\bm{\Phi}| - \frac{\lambda}{2}\left(\Tr(\bm{\Psi}\bm{\Phi}^{-1})\right)\\
\mathcal{L}_{\mathrm{ELBO}} &= \mathbb{E}_{q(\mathbf{z}|\mathbf{x})}\left[\log p(\mathbf{x}|\mathbf{z})\right] + \frac{1}{2} \log {|\tilde{\bm{\Sigma}}|} - \frac{1+\nu}{2} \log{|\bm{\Phi}|} - \frac{\lambda}{2} \Tr\left((\tilde{\bm{\Sigma}} + \tilde{\bm{\mu}}\tilde{\bm{\mu}}^\top)\bm{\Phi}^{-1}\right) - \frac{\lambda}{2}\left(\Tr(\bm{\Psi}\bm{\Phi}^{-1})\right)\\
\mathcal{L}_{\mathrm{ELBO}} &= \mathbb{E}_{q(\mathbf{z}|\mathbf{x})}\left[\log p(\mathbf{x}|\mathbf{z})\right] + \frac{1}{2} \log {|\tilde{\bm{\Sigma}}|}\nonumber\\
&\quad- \frac{1+\nu}{2} \log{|\bm{\Phi}|} - \frac{1+\nu}{2} \Tr\left((\tilde{\bm{\Sigma}} + \tilde{\bm{\mu}}\tilde{\bm{\mu}}^\top)\left({\bm{\Psi} + \mathbf{z}_{i}\mathbf{z}_{i}^\top}\right)^{-1}\right) - \frac{1+\nu}{2}\left(\Tr(\bm{\Psi}\left({\bm{\Psi} + \mathbf{z}_{i}\mathbf{z}_{i}^\top}\right)^{-1})\right)\label{eq:penult}
\end{align}

Eq. (\ref{eq:penult}) and (\ref{eq:bernlike}) can be combined to obtain the final objective.
\begin{align}
\mathcal{L}_{\mathrm{ELBO}} &= \frac{1}{B}\sum_{i=1}^{B}\Bigg(\sum_{j=1}^{D}\left(\mathbf{x}_i|_j\log \hat{\mathbf{x}}_i|_j + (1-\mathbf{x}_i|_j)\log(1 - \hat{\mathbf{x}}_i|_j)\right) + \frac{1}{2} \log {|\tilde{\bm{\Sigma}}|}\nonumber\\
&\quad- \frac{1+\nu}{2} \log{|\left({\bm{\Psi} + \mathbf{z}_{i}\mathbf{z}_{i}^\top}\right)|} - \frac{1+\nu}{2} \Tr\left((\tilde{\bm{\Sigma}} + \tilde{\bm{\mu}}\tilde{\bm{\mu}}^\top)\left({\bm{\Psi} + \mathbf{z}_{i}\mathbf{z}_{i}^\top}\right)^{-1}\right) - \frac{1+\nu}{2}\left(\Tr(\bm{\Psi}\left({\bm{\Psi} + \mathbf{z}_{i}\mathbf{z}_{i}^\top}\right)^{-1})\right)\Bigg)
\end{align}

where $D$ is the dimensionality of input $\mathbf{x}_i$ and $B$ is the batch size.
\end{document}